\documentclass[times, review, 10pt]{elsarticle}
\usepackage{amssymb}
\usepackage{graphicx}
\usepackage{amsmath}
\usepackage{longtable}
\usepackage{a4wide}
\usepackage{mathrsfs}
\usepackage{subcaption}
\usepackage{mathtools}
\usepackage{pifont}
\usepackage{algorithmic}
\usepackage{algorithm}
\usepackage{xcolor}
\usepackage{color}
\usepackage{lineno}
\usepackage{makecell} 
\usepackage{pdflscape}
\usepackage{adjustbox}
\usepackage[utf8]{inputenc}
\usepackage{tabularx}
\usepackage{setspace}
\usepackage{blindtext}
\usepackage{multirow}
\usepackage{notoccite} 
\usepackage{lscape} 
\usepackage{caption} 
\usepackage{mwe}
\usepackage{tikz}
\usepackage{siunitx}
\usepackage{mathrsfs}
\usetikzlibrary{shapes,arrows}
\usepackage{xcolor}
\usepackage{booktabs}
\usepackage{hyperref}
\usepackage{amsthm}

\newtheorem{manualtheoreminner}{Theorem}
\newenvironment{manualtheorem}[1]{%
  \renewcommand\themanualtheoreminner{#1}%
  \manualtheoreminner
}{\endmanualtheoreminner}

\newcommand{\RNum}[1]{\lowercase\expandafter{\romannumeral #1\relax}}
\newcommand{\RNumU}[1]{\uppercase\expandafter{\romannumeral #1\relax}}
\usepackage{natbib}
\journal{Elsevier}

\begin{document}
\date{}
\begin{frontmatter}
\title{Intuitionistic Fuzzy Generalized Eigenvalue Proximal Support Vector Machine}
\author[inst1]{A. Quadir}
\affiliation[inst1]{organization={Department of Mathematics},
            addressline={Indian Institute of Technology Indore}, 
            city={Simrol, Indore},
            postcode={453552}, 
            country={India}}
\author[inst2]{M. A. Ganaie}
\affiliation[inst2]{organization={Department of Computer Science and Engineering},
            addressline={Indian Institute of Technology Ropar}, 
            city={Punjab},
            postcode={140001}, 
            country={India}}
\author[inst1]{M. Tanveer\texorpdfstring{\corref}{corref}{Correspondingauthor}}
 \cortext[Correspondingauthor]{Corresponding author}
\begin{abstract}
Generalized eigenvalue proximal support vector machine (GEPSVM) has attracted widespread attention due to its simple architecture, rapid execution, and commendable performance. GEPSVM gives equal significance to all samples, thereby diminishing its robustness and efficacy when confronted with real-world datasets containing noise and outliers. In order to reduce the impact of noises and outliers, we propose a novel intuitionistic fuzzy generalized eigenvalue proximal support vector machine (IF-GEPSVM). The proposed IF-GEPSVM assigns the intuitionistic fuzzy score to each training sample based on its location and surroundings in the high-dimensional feature space by using a kernel function. The solution of the IF-GEPSVM optimization problem is obtained by solving a generalized eigenvalue problem. Further, we propose an intuitionistic fuzzy improved generalized eigenvalue proximal support vector machine (IF-IGEPSVM) by solving the standard eigenvalue decomposition resulting in simpler optimization problems with less computation cost which leads to an efficient intuitionistic fuzzy-based model. We conduct a comprehensive evaluation of the proposed IF-GEPSVM and IF-IGEPSVM models on UCI and KEEL benchmark datasets. Moreover, to evaluate the robustness of the proposed IF-GEPSVM and IF-IGEPSVM models, label noise is introduced into some UCI and KEEL datasets. The experimental findings showcase the superior generalization performance of the proposed IF-GEPSVM and IF-IGEPSVM models when compared to the existing baseline models, both with and without label noise. Our experimental results, supported by rigorous statistical analyses, confirm the superior generalization abilities of the proposed IF-GEPSVM and IF-IGEPSVM models over the baseline models. Furthermore, we implement the proposed IF-GEPSVM and IF-IGEPSVM models on the USPS recognition dataset, yielding promising results that underscore the models' effectiveness in practical and real-world applications. The source code of the proposed IF-GEPSVM and IF-IGEPSVM models are available at \url{https://github.com/mtanveer1/IF-GEPSVM}.
\end{abstract}
\end{frontmatter}

\section{Introduction}
Support vector machines (SVMs) \cite{cortes1995support} is one of the most successful machine learning tools for classification and regression. SVM is based on statistical learning theory and have been applied to numerous real-world problems such as bio-medicine \cite{noble2004support}, activity recognition \cite{khemchandani2016robust}, image processing \cite{guo2008customizing}, text categorization \cite{joachims2005text} and so on. The main idea of SVM is to seek an optimal plane by maximizing the margin between two parallel supporting hyperplanes. SVM solves one large quadratic programming problem (QPP), resulting in escalated computational complexity, which renders it less suitable for large-scale datasets. It also implements the structural risk minimization (SRM) principle, leading to improved generalization performance. 

Although SVM has made significant strides in various fields, there remains considerable room for improvement. A notable obstacle for the SVM is the substantial computational intricacy involved in solving the QPP. \citet{mangasarian2005multisurface} proposed the generalized eigenvalue proximal SVM (GEPSVM), to mitigate the adverse impact of high computational consumption. GEPSVM aims to find two non-parallel hyperplanes such that each hyperplane is closer to the samples in one class and far away from the samples in the other class by solving two generalized eigenvalue problems. In improved GEPSVM (IGEPSVM) \cite{shao2012improved}, the standard eigenvalue decomposition replaces the generalized eigenvalue decomposition, leading to simpler optimization problems without the potential for singularity. \citet{khemchandani2007twin} proposed twin SVM (TSVM) to solve two smaller-sized QPPs to obtain two non-parallel hyperplanes. Compared with solving one entire QPP in SVM, making TSVM is four times faster than SVM \cite{tanveer2022comprehensive}. \citet{suykens1999least} introduced a variant of SVM known as the least squares support vector machine (LSSVM), to decrease the training cost. LSSVM solves system of linear equations by using a squared loss function instead of the hinge loss. In order to further diminish the training cost, \citet{kumar2009least} introduced a twin variant of LSSVM known as the least square TSVM (LSTSVM). The computation time of LSTSVM is much less in comparison to TSVM. Several modified TSVM models have been proposed, each based on various considerations, including universum TSVM (UTSVM) \cite{qi2012twin}, KNN weighted reduced universum for class imbalance learning \cite{ganaie2022knn}, inverse free reduced universum TSVM for imbalanced data classification (IRUTSVM) \cite{moosaei2023inverse}, and elastic net TSVM and its safe screening rules (SSR-ETSVM) \cite{wang2023elastic}. Other latest advanced models are detailed in \cite{gupta2021kernel, gupta2023least, xie2023deep, pan2023non}.

In real-world environments, the presence of noise and outliers necessitates careful consideration. SVM fails to find an optimal hyperplane when the support vectors are contaminated with noise or outliers, leading to suboptimal or inferior results. To mitigate the influence of noise and outliers, the pinball loss function is employed in SVM (pin-SVM) \cite{huang2013support}. The $\epsilon$-insensitive zone is incorporated into the pin-SVM \cite{huang2013support} to retain the sparsity of the model. Several variants of TSVM have been proposed to mitigate the influence of noise and outliers including $L_1$ norm LSTSVM (NELSTSVM) \cite{gao20111}, general TSVM with pinball loss (pin-GTSVM) \cite{tanveer2019general}, twin parametric margin SVM with pinball loss (Pin-TSVM) \cite{xu2016novel}, ramp loss K-nearest neighbor-weighted multi-class TSVM (RKWMTSVM) \cite{wang2022ramp}, multi-view universum TSVM with insensitive pinball loss (Pin-MvUTSVM) \cite{lou2024multi}, universum TSVM with pinball loss function (Pin-UTSVM) \cite{ganaie2023eeg}, granular ball TSVM with pinball loss (Pin-GBTSVM) \cite{abdul2024granular}, and large scale pinball TSVM (LPTWSVM) \cite{tanveer2022large}. In addition to different loss functions, fuzzy SVM (FSVM) \cite{ganaie2022large, batuwita2010fsvm,lin2002fuzzy} has been proposed, to alleviate the influence of noise and outliers. FSVM calculates the degree of membership function of an input sample based on its individual contribution. This approach enhances the generalization ability of SVMs and mitigates the influence of noise and outliers. Due to superior performance, FSVM has gained popularity in classification tasks and has found broader applications in various domains, including human identification \cite{laxmi2023human}, medical applications \cite{XIAN20106737}, credit risk evaluation \cite{1556587}, and so on. Furthermore, various variants of FSVM have emerged, including a novel fuzzy TSVM \cite{chen2018new} and FSVM for regression estimation \cite{sun2003fuzzy}, which incorporates a novel fuzzy membership function for addressing two-class problems \cite{tang2011fuzzy}. When assigning membership degrees to training points based on their distance from the respective class center, patterns closer to the class center contribute equally to learning the decision surface. However, this approach can lead to certain edge support vectors being incorrectly identified as outliers. Later on, Ha et al. \cite{ming2011intuitionistic} proposed an intuitionistic fuzzy SVM (IF-SVM), in which the influence of each training point on the learning of the decision surface is determined by two parameters: hesitation and degrees of membership. \citet{rezvani2019intuitionistic} introduced intuitionistic fuzzy TSVM (IF-TSVM), by incorporating the degree of membership and non-membership function to further reduce the impact of noise. IF-TSVM has large computational complexity as it solves two QPPs to obtain the optimal hyperplane. Also, IF-TSVM involves matrix inverse computation, which can become impractical in large-scale problems, and may potentially lead to singularity issues. To get motivated by the intuitionistic fuzzy membership scheme and superior performance of GEPSVM, in this paper, we propose intuitionistic fuzzy generalized eigenvalue proximal support vector machine (IF-GEPSVM) by solving the generalized eigenvalue problem in the intuitionistic fuzzy environment. A score, consisting of both membership and non-membership degrees, is assigned to each training instance. The degree of membership function is calculated by the distance between the samples and the corresponding class center while the non-membership functions leverage the statistical correlation between the count of heterogeneous samples to all the samples within their neighborhoods. This membership scheme allows the model to effectively handle outliers and noise that have trespassed in the dataset. 
To address concerns related to the singular value problem and further enhance the speed of obtaining optimal solutions, we propose intuitionistic fuzzy improved generalized eigenvalue proximal support vector machine (IF-IGEPSVM). IF-IGEPSVM solves two standard eigenvalue problems which resolve the singularity issues in IF-GEPSVM. The main highlights of this paper are as follows:
\begin{itemize}
     \item We propose an intuitionistic fuzzy GEPSVM (IF-GEPSVM) and intuitionistic fuzzy IGEPSVM (IF-IGEPSVM). The score value based on the intuitionistic fuzzy number is assigned to each training sample according to their importance in learning the classifier.
     \item We provide rigorous mathematical frameworks for both IF-GEPSVM and IF-IGEPSVM, covering linear and non-linear kernel spaces. Training in the kernel space elevates the proposed models' performance by effectively capturing intricate data patterns and complex relationships through non-linear transformations.
       \item We carried out experiments on artificial datasets and $62$ UCI and KEEL benchmark datasets from diverse domains. The experimental outcomes validate the effectiveness of the proposed IF-GEPSVM and IF-IGEPSVM models when compared to the baseline models.
       \item The proposed IF-GEPSVM and IF-IGEPSVM models undergo rigorous testing with the addition of noise to datasets. Results indicate that the proposed models exhibit robustness to noise and stability to resampling, highlighting their effectiveness under noisy conditions.
       \item As an application, we conducted an experiment on USPS recognition datasets, the numerical experimental demonstrates the superiority of the proposed IF-GEPSVM and IF-IGEPSVM models over the baseline models.
\end{itemize}

This paper is organized as follows: Section \ref{Background} discusses a brief overview of GEPSVM, IGEPSVM, and intuitionistic fuzzy membership schemes.  Section \ref{Proposed work} presents the formulation of the proposed IF-GEPSVM and IF-IGEPSVM models, respectively. Section \ref{Experimental Results} provides a detailed explanation of the experimental results. Finally, the conclusions and potential future research directions are given in Section \ref{Conclusion}.

\section{Background} 
\label{Background}
In this section, we first go through the architecture of GEPSVM and IGEPSVM along with mathematical formulation and intuitionistic fuzzy membership. Let $D=\{(x_i, y_i), i=1,2, \ldots, m\},$ be the traning dataset where $y_i \in \{+1,-1\} $  represents the label of $x_i \in \mathbb{R}^{1 \times n}$. Let us consider the input matrices $A$ $\in \mathbb{R}^{m_{1}\times n}$ and $B$ $\in \mathbb{R}^{m_{2}\times n}$ representing the data points of $+1$ class and $-1$ class respectively, where \( m_1 \)  and \( m_2 \) represent the number of data samples belonging to the $+1$ and $-1$ class, with the total number of data samples being \( m = m_1 + m_2 \). The number of features is denoted by $n$ and $e$ represents the vectors of ones of appropriate dimension.



\subsection{ Rayleigh Quotient}
The Rayleigh quotient \cite{parlett1998symmetric}, denoted as $Q(N,y)$, is defined for a given real symmetric matrix $N \in \mathbb{R}^{m \times m}$ and a nonzero real vector $y \in \mathbb{R}^{m \times 1}$,
\begin{equation}
    Q(N,y)= \frac{y^TNy}{y^Ty}.
\end{equation}
The Rayleigh quotient, $Q(N,y)$, achieves its  maximum (minimum) value, $\lambda_{max}$ ($\lambda_{min}$), when $y$ is equal to $\mu_{max}$ ($\mu_{min}$).
Here, $\mu_{max}$ ($\mu_{min}$) refers to the eigenvector of the generalized eigenvalue problem $Ay = \lambda By$, which corresponds to the maximum (minimum) eigenvalue.

\subsection{Generalized Eigenvalue Proximal Support Vector Machine (GEPSVM)} 
GEPSVM \cite{mangasarian2005multisurface} generates a pair of non-parallel hyperplanes 
\begin{equation}
\label{eq:2}
     w_{1}^T x + b_{1} = 0, \hspace{0.5cm} \text{and} \hspace{0.5cm} w_{2}^Tx + b_{2} = 0,
\end{equation}
such that each hyperplane is closer to the data sample of one class and it is farther from the data samples of another class. The optimization problem of GEPSVM is given by: 
\begin{equation}
\label{eq:3}
    \underset{ (w_{1}, b_{1}) \neq 0}{min} \hspace{0.2cm} \frac{\| A w_{1} + e_{1} b_{1}\|^{2}+\delta\| \tbinom{w_{1}}{b_{1}}\|^{2}}{\|  B w_{1} + e_{2} b_{1}\|^{2}},
\end{equation}
and
\begin{equation}
\label{eq:4}
    \underset{ (w_{2}, b_{2}) \neq 0}{min} \hspace{0.2cm} \frac{\| B w_{2} + e_{2} b_{2}\|^2+\delta\| \tbinom{w_{2}}{b_{2}}\|^{2}}{\| A w_{2} + e_{1} b_{2}\|^2},
\end{equation}
were $\delta > 0$ is a regularization parameter. To make the notation simpler, we introduce $G=[A \hspace{0.2cm} e_{1}]^T [A \hspace{0.2cm} e_{1}] \in \mathbb{R}^{(n+1) \times (n+1)}$, $H=[B \hspace{0.2cm} e_{2}]^T [B \hspace{0.2cm} e_{2}] \in \mathbb{R}^{(n+1) \times (n+1)} $ are symmetric matrices, $z_{1}=[w_{1}; \hspace{0.2cm} b_{1}] \in \mathbb{R}^{(n+1)} $, and $z_{2}=[w_{2}; \hspace{0.2cm} b_{2}]\in \mathbb{R}^{(n+1)}$. By solving the following generalized eigenvalue problems to obtain the solution
\begin{align}
    (G+\delta I)z_{1}=\lambda Hz_{1}, \hspace{0.2cm} z_1 \neq 0,\\
    (H+\delta I)z_{2}=\lambda Gz_{2}, \hspace{0.2cm} z_2 \neq 0,  
\end{align}
where $I$ denotes an identity matrix of appropriate dimension.

Once the optimal values of $z_1$ and $z_2$ are obtained. The classification of a new input data point $x$ into either the class $+1$ or $-1$ can be determined as follows:
\begin{align}
    class(x) =  \underset{i \in \{1, 2\}}{\arg\min} \frac{\lvert w_i^Tx + b_i\rvert}{\|w_i\|}.
\end{align}

\subsection{Improved GEPSVM (IGEPSVM)}
IGEPSVM \cite{shao2012improved} seeks to determine two non-parallel hyperplanes, with each having a small distance from its respective class and a large distance from the other class. IGEPSVM employs subtraction instead of a ratio to learn the non-parallel hyperplanes. The optimization problem of IGEPSVM is as follows:
\begin{equation}
\label{eq:8}
    \underset{ (w_{1}, b_{1}) \neq 0}{min} \hspace{0.2cm} \frac{\| A w_{1} + e_{1} b_{1}\|^2}{\|w_{1}\|^2+b_{1}^2} - \delta \frac{\| B w_{1} + e_{2} b_{1}\|^2}{\|w_{1}\|^2+b_{1}^2},
\end{equation}
and
\begin{equation}
\label{eq:9}
    \underset{ (w_{2}, b_{2}) \neq 0}{min} \hspace{0.2cm} \frac{\| B w_{2} + e_{2} b_{2}\|^2}{\|w_{2}\|^2+b_{2}^2} - \delta \frac{\| A w_{2} + e_{1} b_{2}\|^2}{\|w_{2}\|^2+b_{2}^2},
\end{equation}
where $\delta$ denotes the weight factor. Then the global optimal solution of IGEPSVM can be obtained by solving the following standard eigenvalue problems: 
\begin{align}
    (G+\eta I - \delta H) z_1 = \lambda_1 z_1, \\
    (H+\eta I - \delta G) z_2 = \lambda_2 z_2.\end{align}
After obtaining the optimal values of $z_1$ and $z_2$. The classification of a new input data point $x$ into either the class $+1$ or $-1$ can be determined as follows:
\begin{align}
    class(x) =  \underset{i \in \{1, 2\}}{\arg\min} \frac{\lvert w_i^Tx + b_i\rvert}{\|w_i\|}.
\end{align}

\subsection{Intuitionistic Fuzzy Membership Scheme}
The concept of the fuzzy set was introduced by Zadeh \cite{zadeh1965fuzzy} in 1965, and the intuitionistic fuzzy set (IFS) was subsequently proposed by \citet{atanassov1999intuitionistic} as a means to address uncertainties. It enables a precise representation of situations through the utilization of current information and observations \cite{ha2013support}. In IFS scheme, there are three parameters in an intuitionistic fuzzy number (IFN): the membership degree, non-membership degree, and degree of hesitation, which is denoted as $\nu (0 \leq \nu \leq 1)$, $\mu (0 \leq \mu \leq 1)$ and $\pi = 1 - \nu - \mu$ \cite{rezvani2019intuitionistic}. According to the intuitionistic fuzzy membership (IFM) scheme, every training sample is allocated an IFN, denoted as $(\mu, \nu)$. Finally, a score function is formulated based on $\mu$ and $\nu$ values to examine the presence of outliers within the dataset. The degree of membership and nonmembership function is expressed as follows:

\begin{enumerate}
\item \textit{Membership Function }
The membership function is defined as the distance between the sample and the corresponding class center within the feature space. The membership function of each $i^{th}$ training sample is defined as:

\begin{equation}
    \mu(x_{i}) = 
            \begin{cases}
                  1 - \frac{\|\psi(x_{i}) - C^+\|}{r^+ + \hspace{0.2cm} \gamma}, & y_{i} = +1,\\
                   1 - \frac{\|\psi(x_{i}) - C^-\|}{r^- + \hspace{0.2cm} \gamma}, & y_{i} = -1,
            \end{cases}
\end{equation}
where $\gamma$ is a non-negative parameter, $\psi$ represents the feature (projection) mapping function, and $r^+ (r^-)$ are the radius of $+1$ $(-1)$ class given by 
\begin{align}
    r^{+}=\underset{y_{i}= +1}\max \Vert \ \psi(x_{i})-C^{+}\Vert, \hspace{0.2cm} \text{and} \hspace{0.2cm} r^{-}=\underset{y_{i}= -1}\max \Vert \ \psi(x_{i})-C^{-}\Vert,
\end{align}
where $C^+ (C^-)$ are the center of $+1$ $(-1)$ class, respectively. The class center is defined as
\begin{align}
    C^+ = \frac{1}{m_{1}} \sum _{y_{i}= +1} \psi(x_{i}), \hspace{0.2cm} \text{and} \hspace{0.2cm} C^- = \frac{1}{m_{2}} \sum _{y_{i}= -1} \psi(x_{i}),
\end{align}
here $m_{1}$ and $m_{2}$ are the number of $+1$ and $-1$ class samples, respectively.

\item \textit{Non-membership function}
The non-membership function for each training sample is defined as the proportion of heterogeneous points to the total number of points within its vicinity. Therefore, the nonmembership function is calculated as follows:
\begin{equation}
    \nu(x_{i}) = (1 - \mu (x_{i})) \eta (x_{i}),
\end{equation}
where the local neighborhood set $\eta(x_i)$ is calculated as:
\begin{equation} 
\eta (x_{i})=\frac{|\lbrace x_{j}|\Vert  \psi(x_{i})- \psi(x_{j})\Vert \leq \beta,\,y_{j}\ne y_{i}\rbrace |}{|\lbrace x_{j}|\Vert  \psi(x_{i})- \psi(x_{j})\Vert \leq \beta \rbrace |},
\end{equation}
here $\beta$ is an adjustable non-negative parameter and $\lvert \cdot \rvert$ represents the cardinality.

\item \textit{Score of each training sample:}
The score function amalgamates the significance of membership and non-membership, and this can be calculated as:

\begin{equation}
\label{eq:18}
s_{i}=\left\lbrace \begin{array}{lr} \mu _{i}, & \nu _{i}=0,\\ 0, & \mu _{i}\leq \nu _{i}, \\ \frac{1-\nu _{i}}{2-\mu _{i}-\nu _{i}}, & \text{others}. \end{array}\right. 
\end{equation}
Finally, the score matrix \( S \) for the dataset \( D \) is defined as: $S=diag\{s(x_i) \hspace{0.1cm} | \hspace{0.1cm} i=1,2, \ldots, m\}$.
\end{enumerate}

We calculate the intuitionistic fuzzy membership scheme in high-dimensional space. It is essential to accurately compute the distance between data points in this high-dimensional feature space. This is where the kernel function becomes crucial, as it allows us to implicitly map data points into a higher-dimensional space without needing an explicit formula for the transformation. Specifically, the kernel function $k(x_i, x_j) = \psi(x_i)^T\psi(x_j)$ defines the inner product between the images of the data points $x_i$ and $x_j$ in the feature space. However, since we do not have an explicit representation of $\psi(x)$, calculating the direct distance between points is not straightforward. Theorems \ref{Th1} and \ref{Th2} provide a mathematical framework to compute the distances between the points and the distance between the point to the center in the kernel space effectively, ensuring that the intuitionistic fuzzy membership values are calculated accurately. Therefore, these theorems are not just for structure but are essential components that leverage the kernel function to calculate the intuitionistic fuzzy membership scheme in high-dimensional spaces accurately.

\begin{manualtheorem}{1}
\label{Th1}
\cite{ha2013support}: Let the kernel function is $K(x_{i}, x_j)$. Then, the inner product distance is given by:\\
$\|\psi(x_i) -\psi(x_j)\|=\sqrt{K(x_{i}, x_i) - 2K(x_{i}, x_j) + K(x_{j}, x_j)}$.
\end{manualtheorem}
\begin{proof}
\begin{align}
         \|\psi(x_i) -\psi(x_j)\|&=\sqrt{(\psi(x_i) -\psi(x_j)).(\psi(x_i) -\psi(x_j))} \nonumber \\
            & = \sqrt{(\psi(x_i).\psi(x_i)) - (\psi(x_i).\psi(x_j)) + (\psi(x_j).\psi(x_j))}  \nonumber \\
            & = \sqrt{K(x_{i}, x_i) - 2K(x_{i}, x_j) + K(x_{j}, x_j)}. \nonumber
\end{align}
\end{proof}

\begin{manualtheorem}{2}
\label{Th2}
   \cite{ha2013support}: The Euclidean distance between the samples and the corresponding class center is represented by: \\
    $\|\psi(x_i) - C^+\|=\sqrt{K(x_{i}, x_i) - \frac{2}{m_1}\sum_{y_j = +1}K(x_{i}, x_j) + \frac{1}{m_1^2}\sum_{y_i = +1}\sum_{y_j = +1}K(x_{i}, x_j)}$, \\
    $\|\psi(x_i) - C^-\| = \sqrt{K(x_{i}, x_i) - \frac{2}{m_2}\sum_{y_j = -1}K(x_{i}, x_j) + \frac{1}{m_2^2}\sum_{y_i = -1}\sum_{y_j = -1}K(x_{i}, x_j)}$.
\end{manualtheorem}
\begin{proof}
\begin{align}
    \|\psi(x_i) - C^+\|&=\sqrt{(\psi(x_i) - C^+).(\psi(x_i) - C^+)}  \nonumber \\
    & = \sqrt{(\psi(x_i).\psi(x_i)) + (C^+.C^+) -  2(\psi(x_i).C^+)}  \nonumber \\
    & = \sqrt{K(x_{i}, x_i) +  \left (\frac{1}{m_1}\sum_{y_i = +1}\psi(x_i)\right). \left (\frac{1}{m_1}\sum_{y_i = +1}\psi(x_i)\right)  -  2\psi(x_i) \left (\frac{1}{m_1}\sum_{y_j = +1}\psi(x_i)\right)} \nonumber \\
    & = \sqrt{K(x_{i}, x_i) +  \frac{1}{m_1^2}\sum_{y_i = +1}\sum_{y_j = +1}K(x_{i}, x_j)  -  \frac{2}{m_1}\sum_{y_j = +1}K(x_{i}, x_j)}. \nonumber
\end{align}
Similarly, $\|\psi(x_i) - C^-\|$ can be calculated.
\end{proof}

\section{Proposed work}
\label{Proposed work}
In traditional machine learning models, including GEPSVM and IGEPSVM, each data sample is given the same weight irrespective of its nature. Inherent to datasets in the presence of noise and outliers, making their occurrence a natural phenomenon. While the presence of noise and outliers is inherent and expected, their effect on traditional GEPSVM and IGEPSVM models is detrimental. Hence, to address the presence of noisy samples and outliers within the dataset, we propose an intuitionistic fuzzy generalized eigenvalue proximal support vector machine (IF-GEPSVM) and intuitionistic fuzzy improved generalized eigenvalue proximal support vector machine (IF-IGEPSVM). In the proposed IF-GEPSVM and IF-IGEPSVM models, the intuitionistic fuzzy membership value is determined by the proximity of a sample to the class center in the high-dimensional feature space, respectively. The fuzzy membership values quantify the extent to which a sample is associated with a particular class. The fuzzy non-membership value in the IF-GEPSVM and IF-IGEPSVM models is calculated by taking into account the neighborhood information of the sample, indicating the degree to which it does not belong to a specific class.

\subsection{Intuitionistic Fuzzy Generalized Eigenvalue Proximal Support Vector Machine (IF-GEPSVM)}
In this subsection, we provide a detailed mathematical formulation of the proposed IF-GEPSVM model tailored for linear and non-linear cases.

\subsubsection{Linear IF-GEPSVM}
The optimization problem of the proposed IF-GEPSVM model for the linear case is defined as follows:
\begin{equation}
\label{eq:19}
    \underset{ (w_{1}, b_{1}) \neq 0}{min} \hspace{0.2cm} \frac{\|S_{1}[ Aw_{1} + e_{1} b_{1}]\|^2+\delta\| \tbinom{w_{1}}{b_{1}}\|^{2}}{\|S_{2}[ Bw_{1} + e_{2} b_{1}]\|^2},
\end{equation}
and
\begin{equation}
\label{eq:20}
    \underset{ (w_{2}, b_{2}) \neq 0}{min} \hspace{0.2cm} \frac{\|S_{2}[ Bw_{2} + e_{2} b_{2}]\|^2+\delta\| \tbinom{w_{2}}{b_{2}}\|^{2}}{\|S_{1}[ Aw_{2} + e_{1} b_{2}]\|^2},
\end{equation}
where $S_1=\text{diag}(s_i), \hspace{0.2cm} \forall i= 1,2 \hdots m_1$, and $S_2=\text{diag}(s_j), \hspace{0.2cm} \forall j= 1,2 \hdots m_2$, are the score matrix for the dataset of $+1$ and $-1$ class, respectively. The numerator term in the problem \eqref{eq:19} minimizes the distance between the positive hyperplane to samples in the $+1$ class. The denominator term maximizes the distance of the positive hyperplane from the samples in $-1$ class. Similarly, we can draw the inference of the problem \eqref{eq:20}. To simplify the notation, we introduce $G=[S_{1}(A, \hspace{0.1cm} e_{1})]^T [S_{1}(A, \hspace{0.1cm} e_{1})] \in \mathbb{R}^{(n+1) \times (n+1)}$ and $H=[S_{2}(B, \hspace{0.1cm} e_{2})]^T [S_{2}(B, \hspace{0.1cm} e_{2})] \in \mathbb{R}^{(n+1) \times (n+1)}$ are symmetric matrices.

Then the optimization problems represented by $\eqref{eq:19}$ and $\eqref{eq:20}$ can be reformulated as:
\begin{align}
\label{eq:21}
    \underset{ z_{1}\neq 0}{min} \hspace{0.2cm} \frac{z_{1}^T(G + \delta I)z_{1}}{z_{1}^THz_{1}},
\end{align}
and 
\begin{align}
\label{eq:22}
    \underset{ z_{2} \neq 0}{min} \hspace{0.2cm} \frac{z_{2}^T(H + \delta I)z_{2}}{z_{2}^TGz_{2}}.
\end{align}
Problems \eqref{eq:21} and \eqref{eq:22} can be classified as generalized Rayleigh quotients; therefore, by solving the generalized eigenvalue problems to obtain the optimal solution:
\begin{align}
    (G+\delta I)z_{1}=\lambda Hz_{1}, \hspace{0.2cm} z_1 \neq 0, \label{eq:23}\\
    (H+\delta I)z_{2}=\lambda Gz_{2}, \hspace{0.2cm} z_2 \neq 0, \label{eq:24}  
\end{align}
where $I$ denotes an identity matrix of appropriate dimension.

The optimal parameters $z_1$ and $z_2$ of the hyperplane are obtained by the eigenvector corresponding to the smallest eigenvalues. The classification of a new input data point $x$ can be determined as follows:
\begin{align}
\label{eq:25}
    class(x) =  \underset{i \in \{1, 2\}}{\arg\min} \frac{\lvert w_i^Tx + b_i\rvert}{\|w_i\|}.
\end{align}

\subsubsection{Non-linear IF-GEPSVM}
The performance of linear classifiers falls dramatically once samples are linearly nonseparable in the input space.  In this subsection, we extend the linear IF-GEPSVM model to the non-linear case. IF-GEPSVM with non-linear kernel finds the hyperplanes given by
\begin{align}
    w_1^T K(x, C^T) + b_1 = 0 \hspace{0.2cm} \text{and} \hspace{0.2cm} w_2^T K(x, C^T) + b_2 = 0, 
\end{align}
where $K$ is a kernel function and $C=[A;\hspace{0.1cm} B]$.

The optimization problem of IF-GEPSVM for non-linear case is as follows:

\begin{equation}
\label{eq:27}
    \underset{ (w_{1}, b_{1}) \neq 0}{min} \hspace{0.2cm} \frac{\|S_{1}[ K(A,\hspace{0.1cm} C^T)w_{1} + e_{1} b_{1}]\|^2+\delta\| \tbinom{w_{1}}{b_{1}}\|^{2}}{\|S_{2}[ K(B,\hspace{0.1cm} C^T)w_{1}  + e_{2} b_{1}]\|^2},
\end{equation}
and
\begin{equation}
\label{eq:28}
    \underset{ (w_{2}, b_{2}) \neq 0}{min} \hspace{0.2cm} \frac{\|S_{2}[ K(B,\hspace{0.1cm} C^T)w_{2} + e_{2} b_{2}]\|^2+\delta\| \tbinom{w_{2}}{b_{2}}\|^{2}}{\|S_{1}[ K(A,\hspace{0.1cm} C^T)w_{2}  + e_{1} b_{2}]\|^2},
\end{equation}
where $\delta$ is a weighting factor. To make the notation simpler, we introduce $P_1=[S_{1}K(A,\hspace{0.1cm} C^T), \hspace{0.1cm} S_{1}e_{1}]$ and $Q_1=[S_{2}K(B,\hspace{0.1cm} C^T), \hspace{0.1cm} S_{2}e_{2}]$.

The optimization problem \eqref{eq:27} and \eqref{eq:28} becomes
\begin{equation}
\label{eq:29}
    \underset{ z_{1}\neq 0}{min} \hspace{0.2cm} \frac{z_{1}^T(P + \delta I)z_{1}}{z_{1}^TQz_{1}}, \hspace{1cm} \text{and} \hspace{1cm} \underset{ z_{2} \neq 0}{min} \hspace{0.2cm} \frac{z_{2}^T(Q + \delta I)z_{2}}{z_{2}^TPz_{2}}.
\end{equation}

By solving the generalized eigenvalue problem to obtain the global optimal solution 
\begin{align}
    (P+\delta I)z_{1}=\lambda Qz_{1}, \hspace{0.2cm} z_1 \neq 0, \hspace{0.2cm} \text{and} \hspace{0.2cm}
    (Q+\delta I)z_{2}=\lambda Pz_{2}, \hspace{0.2cm} z_2 \neq 0,  
\end{align}
The classification of a new input data point $x$ into either the $+1$ or $-1$ can be determined as follows:
\begin{equation}
    class(x) =  \hspace{0.2cm} \underset{ i=\{1,2\} }{\arg\min} \hspace{0.2cm} \frac{\lvert K(x^T, \hspace{0.2cm} C^T)w_{i}+b_{i} \rvert}{\sqrt{w_{i}^TK(C, \hspace{0.2cm} C^T)w_{i}}}.
\end{equation}

\subsection{Intuitionistic Fuzzy Improved Generalization Eigenvalue Proximal Support Vector Machine (IF-IGEPSVM)}
IF-GEPSVM still poses risks in certain cases, for instance, the issue of singularity is prone to occur during the implementation of generalized eigenvalue decomposition. In order to address the limitations of IF-GEPSVM and reduce the training time. We present an intuitionistic fuzzy improved generalization eigenvalue proximal support vector machine (IF-IGEPSVM) to mitigate the impact of noise and outliers.

\subsubsection{Linear IF-IGEPSVM}
The optimization problem of IF-IGEPSVM for linear case is defined as follows:
\begin{equation}
    \underset{ (w_{1}, b_{1}) \neq 0}{min} \hspace{0.2cm} \frac{\|S_{1}[ Aw_{1} + e_{1} b_{1}]\|^2}{\|w_{1}\|^2+b_{1}^2} - \delta \frac{\|S_{2}[ Bw_{1} + e_{2} b_{1}]\|^2}{\|w_{1}\|^2+b_{1}^2},
\end{equation}
and
\begin{equation}
    \underset{ (w_{2}, b_{2}) \neq 0}{min} \hspace{0.2cm} \frac{\|S_{2}[ Bw_{2} + e_{2} b_{2}]\|^2}{\|w_{2}\|^2+b_{2}^2} - \delta \frac{\|S_{1}[ Aw_{2} + e_{1} b_{2}]\|^2}{\|w_{2}\|^2+b_{2}^2},
\end{equation}
where $\delta > 0$ is a tunable parameter. The above problem is then reduced as follows:

\begin{equation*}
    \underset{ z_{1}, \kappa_{1}}{min} \hspace{0.2cm} \frac{1}{\kappa_{1}}z_{1}^TGz_{1} - \delta \frac{1}{\kappa_{1}}z_{1}^THz_{1} \\
\end{equation*}
\begin{equation}
\label{eq:31}
    s.t. \hspace{0.2cm} \|z_{1}\|^2 = \kappa_{1}, \hspace{0.2cm} \kappa_{1} > 0, 
\end{equation}
and
\begin{equation*}
    \underset{ z_{2}, \kappa_{2} }{min} \hspace{0.2cm} \frac{1}{\kappa_{2}}z_{2}^THz_{2} - \delta \frac{1}{\kappa_{2}}z_{2}^TGz_{2} \\ 
\end{equation*}
\begin{equation}
\label{eq:32}
    s.t. \hspace{0.2cm} \|z_{2}\|^2 = \kappa_{2}, \hspace{0.2cm} \kappa_{2} > 0,
\end{equation}
where $G$ and $H$ are the same as defined above.

By introducing the Tikhonov regularization term, then the problem \eqref{eq:31} and \eqref{eq:32} are given as follows:

\begin{equation*}
    \underset{ (w_{1}, b_{1}) \neq 0}{min} \hspace{0.2cm} z_{1}^TGz_{1}+\eta\|z_{1}\|^{2} - \delta z_{1}^THz_{1}
\end{equation*}
\begin{equation}
\label{eq:35}
   s.t. \hspace{0.2cm} \|z_{1}\|^2 = \kappa_{1}, \hspace{0.2cm} \kappa_{1} > 0,
\end{equation}
and
\begin{equation*}
    \underset{ (w_{2}, b_{2}) \neq 0}{min} \hspace{0.2cm} z_{2}^THz_{2}+\eta\|z_{2}\|^{2} - \delta z_{2}^TGz_{2}
\end{equation*}
\begin{equation}
\label{eq:36}
   s.t. \hspace{0.2cm} \|z_{2}\|^2 = \kappa_{2}, \hspace{0.2cm} \kappa_{2} > 0,
\end{equation}
where $\eta$ is a non-negative parameter.  

Using the Lagrange function for solving $\eqref{eq:35}$, we get
\begin{equation}
    L(z_{1}, \alpha_{1}) = z_{1}^TGz_{1} + \eta \|z_{1}\|^2 - \delta z_{1}^THz_{1} - \alpha_{1}(\|z_{1}\|^2-\kappa_{1}) - \alpha_{2} \kappa_{1},
\end{equation}
where $\alpha_{1}$ and  $\alpha_{2}$ are the Lagrange multipliers. By setting the gradient w.r.t $z_1$ equal to $0$, we get

\begin{equation}
\label{eq:38}
    2(G^T + \eta I)z_{1} - 2\delta H^Tz_{1} - 2\alpha_{1} z_{1} = 0,
\end{equation}
where $I$ is the identity matrix of the appropriate dimension. 

The optimal solution of \eqref{eq:35} is obtained by solving the standard eigenvalue problem:
\begin{equation}
\label{eq:39}
    (G^T+\eta I-\delta H^T)z_{1} = \alpha_{1} z_{1}.
\end{equation}

In a similar way, the solution of \eqref{eq:36} can be obtained by solving standard eigenvalue problem:

\begin{equation}
\label{eq:40}
    (H^T+\eta I-\delta G^T)z_{2} = \alpha_{2} z_{2}.
\end{equation}

The classification of a new input data point $x$ into either the $+1$ or $-1$ can be determined as follows:
\begin{equation}
\label{eq:142}
    Class (x) =  \hspace{0.1cm} \underset{ i=\{1,2\} }{\arg\min} \hspace{0.2cm} \frac{\lvert w_{i}^Tx+b_{i} \rvert}{\|w_{i}\|}.
\end{equation}

\subsubsection{Non-linear IF-IGEPSVM}
In this subsection, we extend the linear IF-IGEPSVM model to the non-linear case by introducing a kernel function. The optimization problem of IF-IGEPSVM is defined as follows:

\begin{equation}
\label{eq:42}
    \underset{ (w_{1}, b_{1}) \neq 0}{min} \hspace{0.1cm} \frac{\|S_{1}[K(A,\hspace{0.1cm} C^T)w_{1} + e_{1} b_{1}]\|^2}{\|w_{1}\|^2+b_{1}^2} - \delta \frac{\|S_{2}[K(B,\hspace{0.1cm} C^T)w_{1} + e_{2} b_{1}]\|^2}{\|w_{1}\|^2+b_{1}^2},
\end{equation}
and
\begin{equation}
\label{eq:43}
    \underset{ (w_{2}, b_{2}) \neq 0}{min} \hspace{0.1cm} \frac{\|S_{2}[K(B,\hspace{0.1cm} C^T)w_{2} + e_{2} b_{2}]\|^2}{\|w_{2}\|^2+b_{2}^2} - \delta \frac{\|S_{1}[K(A,\hspace{0.1cm} C^T)w_{2} + e_{1} b_{2}]\|^2}{\|w_{2}\|^2+b_{2}^2},
\end{equation}
where $\delta$ is a weighting factor, $C=[A;\hspace{0.1cm} B]$ and $K$ is the kernel function.

The problem \eqref{eq:42} and \eqref{eq:43} become

\begin{equation*}
    \underset{ (w_{1}, b_{1}) \neq 0}{min} \hspace{0.2cm} z_{1}^TPz_{1}+\eta\|z_{1}\|^{2} - \delta z_{1}^TQz_{1}
\end{equation*}
\begin{equation}
\label{eq:44}
   s.t. \hspace{0.2cm} \|z_{1}\|^2 = \kappa_{1}, \hspace{0.2cm} \kappa_{1} > 0,
\end{equation}
and
\begin{equation*}
    \underset{ (w_{2}, b_{2}) \neq 0}{min} \hspace{0.2cm} z_{2}^TQz_{2}+\eta\|z_{2}\|^{2} - \delta z_{2}^TPz_{2}
\end{equation*}
\begin{equation}
\label{eq:45}
   s.t. \hspace{0.2cm} \|z_{2}\|^2 = \kappa_{2}, \hspace{0.2cm} \kappa_{2} > 0,
\end{equation}
where $\eta$ is a nonnegative parameter and $P$ and $Q$ are same as defined above.

Likewise in linear case, eigenvalue problems can be used to compute the solution of $\eqref{eq:44}$ and $\eqref{eq:45}$ as follows:
\begin{equation}
    (P^T+\eta I - \delta Q^T)z_{1} = \alpha_{1}z_{1},
\end{equation}
\begin{equation}
    (Q^T+\eta I - \delta P^T)z_{2} = \alpha_{2}z_{2}.
\end{equation}
After obtaining the solution of $(w_{1}, \hspace{0.2cm} b_{1})$ and $(w_{2}, \hspace{0.2cm} b_{2})$ of $\eqref{eq:42}$ and $\eqref{eq:43}$, the class $i \hspace{0.2cm} (i=1, 2)$  is assigned to a new point with respect to the closeness of the two hyperplanes, $i.e.,$ 
 
\begin{equation}
    class(x) =  \hspace{0.1cm} \underset{ i=\{1,2\} }{\arg\min} \hspace{0.1cm} \frac{\lvert K(x^T, \hspace{0.1cm} C^T)w_{i}+b_{i} \rvert}{\sqrt{w_{i}^TK(C, \hspace{0.1cm} C^T)w_{i}}}.
\end{equation}

\section{Discussion of the proposed IF-GEPSVM and IF-IGEPSVM models w.r.t. the baselines models}
In this section, we elucidate the comparison of the proposed IF-GEPSVM and IF-IGEPSVM models and the existing models.
\begin{enumerate}
    \item Difference between Pin-GTSVM and the proposed IF-GEPSVM and IF-IGEPSVM models:
        \begin{itemize}
            \item The proposed IF-GEPSVM and IF-IGEPSVM models solve a pair of eigenvalue problems to find optimal parameters, whereas Pin-GTSVM solves two quadratic programming problems (QPPs) to determine optimal hyperplanes. As a result, for large datasets with numerous features, the computational complexity of Pin-GTSVM typically scales with the size of the input data.
            \item The proposed IF-GEPSVM and IF-IGEPSVM models incorporate intuitionistic fuzzy theory to mitigate the impact of noise and outliers. In contrast, for Pin-GTSVM, the pinball loss function can be more sensitive to outliers and noise, particularly for extreme quantiles (very high or very low). This sensitivity can result in unstable quantile estimates.
            \item The performance of models using pinball loss can be sensitive to the choice of the quantile parameter, which requires careful tuning and may not be straightforward. In contrast, our proposed models find the membership value in high-dimensional space, making them more suitable.
        \end{itemize}
        \item Difference between GEPSVM and IGEPSVM with the proposed IF-GEPSVM and IF-IGEPSVM models:
        \begin{itemize}
            \item The proposed IF-GEPSVM and IF-IGEPSVM models integrate intuitionistic fuzzy theory into the traditional GEPSVM and IGEPSVM framework. Intuitionistic fuzzy sets are characterized by a membership function, a non-membership function, and a degree of hesitation, providing a richer representation of uncertainty.
            \item Our proposed IF-GEPSVM and IF-IGEPSVM models mitigate the impact of noise and outliers by assigning less importance to uncertain or ambiguous data points. This approach results in more stable and reliable eigenvalue computations. In contrast, GEPSVM and IGEPSVM fail to deal with the noise and outliers issues present in the datasets.
        \end{itemize}
        \item Difference between CGFTSVM-ID and the proposed IF-GEPSVM and IF-IGEPSVM models:
        \begin{itemize}
            \item Our proposed IF-GEPSVM and IF-IGEPSVM models solve the eigenvalue problem to obtain the optimal hyperplane. In contrast, implementing the generalized bell fuzzy membership function in the CGFTSVM-ID model may introduce additional computational overhead, particularly during the training and inference phases. This overhead can affect scalability, especially for large-scale datasets or real-time applications.
            \item The proposed IF-GEPSVM and IF-IGEPSVM models assign each training instance a score based on both membership and non-membership degrees. The membership degree assesses the distance of the sample from the class center, while the non-membership degree quantifies the ratio of the heterogeneous sample to the total samples in its neighborhood. This mechanism effectively mitigates the impact of noise and outliers. Conversely, CGFTSVM-ID employs the Generalized Bell Fuzzy scheme, which may be sensitive to noise and outliers in the data. This sensitivity can distort the shape and effectiveness of the membership function, potentially resulting in suboptimal classification outcomes.
            \item The generalized bell fuzzy membership function often requires parameter tuning to achieve optimal performance. This process can be challenging and may necessitate extensive experimentation, particularly with complex datasets. In contrast, our proposed models determine membership values directly in high-dimensional space, which enhances their suitability and efficacy.
        \end{itemize}
\end{enumerate}

\section{Computational Complexity}
Let $m$ denote the total number of training samples, with $p = \frac{m}{2}$ representing the number of samples present in each class. In computing the degree of membership, the proposed IF-GEPSVM model entails the computation of the class radius, the computation of the class center, and determining the distance of each sample from the class center. Therefore, the complexity for determining the membership degree is $\mathcal{O}(1)+\mathcal{O}(1)+\mathcal{O}(p)+\mathcal{O}(p)$. For measuring the degree of non-membership, the computational complexity is $\mathcal{O}(p)+\mathcal{O}(p)$. Hence, the proposed IF-GEPSVM model utilizes $2 \times \mathcal{O}(p)$ operations for assigning the score values. 
The proposed IF-GEPSVM involves solving two generalized eigenvalue problems, with a computational complexity of $\mathcal{O}(n^3)$. Hence the overall computational complexity of the proposed IF-GEPSVM model is $\mathcal{O}(p) + \mathcal{O}(n^3)$ for the linear case. Also, the computational complexity of the proposed IF-IGEPSVM model is $\mathcal{O}(p) + \mathcal{O}(n^2)$ for the linear case. The algorithm of the proposed IF-GEPSVM and IF-IGEPSVM models are briefly described in Algorithm \ref{IF-GEPSVM classifier} and Algorithm \ref{IF-IGEPSVM classifier}, respectively.

\begin{algorithm}
\caption{IF-GEPSVM}
\label{IF-GEPSVM classifier}
\textbf{Input:} $A \in \mathbb{R}^{m_1 \times n}$, and $B \in \mathbb{R}^{m_2 \times n}$ are the matrices of training samples of $+1$ and $-1$ class.\\
\textbf{Output:} IF-GEPSVM model.
\begin{algorithmic}[1]
\STATE Compute score matrices $S_1$ and $S_2$ for samples belonging to class $A$ and $B$ using \eqref{eq:18}, respectively.
\STATE Obtain $G=[S_{1}(A, \hspace{0.1cm} e_{1})]^T [S_{1}(A, \hspace{0.1cm} e_{1})]$ and $H=[S_{2}(B, \hspace{0.1cm} e_{2})]^T [S_{2}(B, \hspace{0.1cm} e_{2})]$,  where \( e_1 \) and \( e_2 \) are column vectors of ones with appropriate dimensions.
\STATE Select the best parameters by using the grid search method.
\STATE Obtain the optimal hyperplanes $(w_1, b_1)$ and $(w_2, b_2)$ for each class by solving the generalized eigenvalue problems \eqref{eq:23} and \eqref{eq:24}. 
\STATE For classifying testing point $x_i$, if $\arg\min \frac{\lvert w_{i}^Tx+b_{i} \rvert}{\|w_{i}\|} =1$, then the data point belongs to class $+1$, otherwise the data point belongs to class $-1$.
\end{algorithmic}
\end{algorithm}

\begin{algorithm}
\caption{IF-IGEPSVM}
\label{IF-IGEPSVM classifier}
\textbf{Input:} $A \in \mathbb{R}^{m_1 \times n}$, and $B \in \mathbb{R}^{m_2 \times n}$ are the matrices of training samples of $+1$ and $-1$ class.\\
\textbf{Output:} IF-IGEPSVM model.
\begin{algorithmic}[1]
\STATE Compute score matrices $S_1$ and $S_2$ for samples belonging to class $A$ and $B$ using \eqref{eq:18}, respectively.
\STATE Obtain $G=[S_{1}(A, \hspace{0.1cm} e_{1})]^T [S_{1}(A, \hspace{0.1cm} e_{1})]$ and $H=[S_{2}(B, \hspace{0.1cm} e_{2})]^T [S_{2}(B, \hspace{0.1cm} e_{2})]$,  where \( e_1 \) and \( e_2 \) are column vectors of ones with appropriate dimensions.
\STATE Select the best parameters by using the grid search method.
\STATE Obtain the optimal hyperplanes $(w_1, b_1)$ and $(w_2, b_2)$ for each class by solving the standard eigenvalue problems \eqref{eq:39} and \eqref{eq:40}. 
\STATE For classifying testing point $x_i$, if $\arg\min \frac{\lvert w_{i}^Tx+b_{i} \rvert}{\|w_{i}\|} =1$, then the data point belongs to class $+1$, otherwise the data point belongs to class $-1$.
\end{algorithmic}
\end{algorithm}

\section{Experimental Results} 
\label{Experimental Results}

This section presents detailed information on the experimental setup, including datasets and compared models. We then analyze the experimental results and perform statistical analyses. We introduce label noise at varying levels, specifically $5\%$, $10\%$, $15\%$, and $20\%$, into each dataset. This involves shifting the labels of a certain percentage of data points from one class to another at these specified levels. We examine the impact of this noise on the performance of the proposed IF-GEPSVM and IF-IGEPSVM models. 

\subsection{ Experimental setup}
To test the efficiency of proposed  IF-GEPSVM and IF-IGEPSVM models, we compare them to baseline models, namely Pin-GTSVM \cite{tanveer2019general}, GEPSVM \cite{mangasarian2005multisurface} IGEPSVM \cite{shao2012improved}, and CGFTSVM-ID \cite{anuradha2024} on publicly available UCI \cite{dua2017uci} and KEEL \cite{derrac2015keel} datasets. Furthermore, we conducted experiments on artificially generated datasets and USPS \footnote{\url{https://cs.nyu.edu/~roweis/data.html}} recognition datasets. The performance of the model is evaluated using MATLAB R$2022$b installed on the Windows $10$ operating system, Intel(R) Xeon(R) Gold $6226$R CPU $@$ $2.90$GHz  CPU and $128$ GB RAM. For the non-linear case, we use Gaussian kernel given by $K(x_i,x_j)=\exp{\left(-\frac{\|x_i-x_j\|^2}{2\sigma^2}\right)}$, here $\sigma$ is the Kernel parameter. The entire data set is divided into a ratio of $70\colon30$ for training and testing, respectively. We used $10$-fold cross-validation to obtain the best hyperparameter tuning for getting better accuracy (ACC). The hyperparameters  of models $c_1, c_2, \delta, \eta, \sigma $ are chosen from the ranges: $\{2^{i} \hspace{0.1cm} | \hspace{0.1cm} i=-8, -7, \ldots, 7, 8\}$. For Pin-GTSVM, to reduce the computational cost of the model, we set $\tau_1 = \tau_2 = 0.5$. For CGFTSVM-ID, hyperparameter $r$ is selected from the range $[0.5, \hspace{0.1 cm} 0.625 ,\hspace{0.1 cm} 0.75, \hspace{0.1 cm} 0.875, \hspace{0.1 cm} 1]$.

\subsection{Experiments on Artificial data}
\begin{figure*}[ht]
\begin{minipage}{.30\linewidth}
\centering
\subfloat[Pin-GTSVM]{\label{fig1:1a}\includegraphics[scale=0.25]{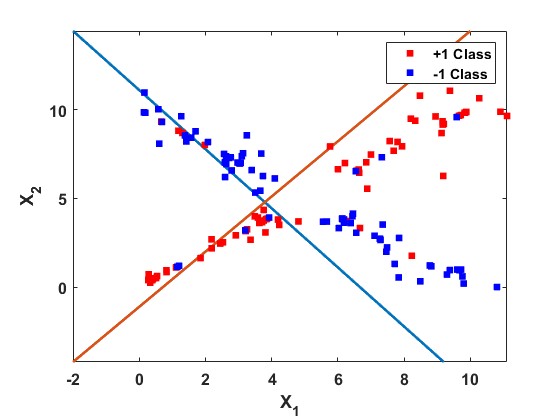}}
\end{minipage}
\begin{minipage}{.30\linewidth}
\centering
\subfloat[GEPSVM]{\label{fig1:1b}\includegraphics[scale=0.25]{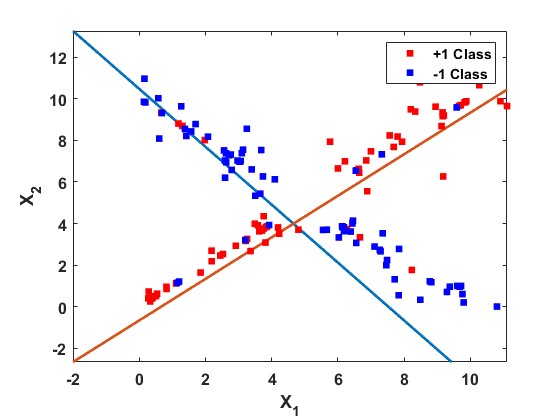}}
\end{minipage}
\begin{minipage}{.30\linewidth}
\centering
\subfloat[IGEPSVM]{\label{fig1:1c}\includegraphics[scale=0.25]{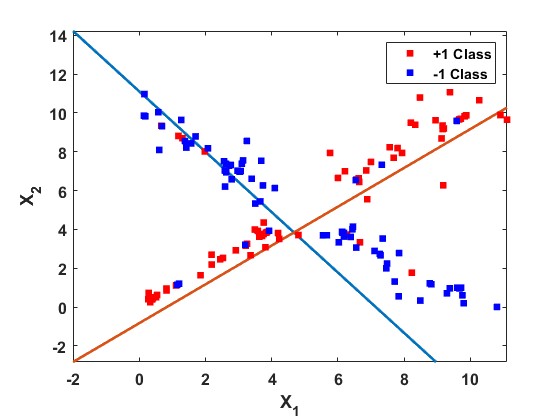}}
\end{minipage}
\par\medskip
\begin{minipage}{.30\linewidth}
\centering
\subfloat[CGFTSVM-ID]{\label{fig1:1d}\includegraphics[scale=0.25]{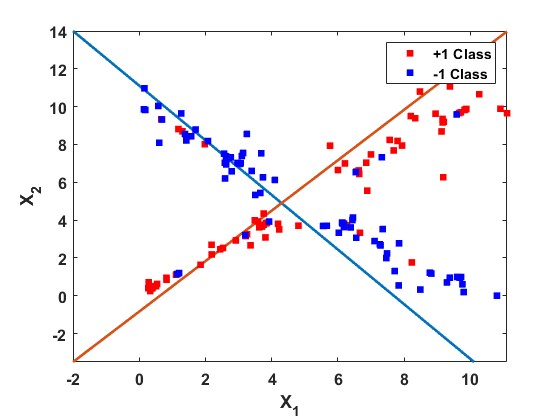}}
\end{minipage}
\begin{minipage}{.30\linewidth}
\centering
\subfloat[IF-GEPSVM]{\label{fig1:1e}\includegraphics[scale=0.25]{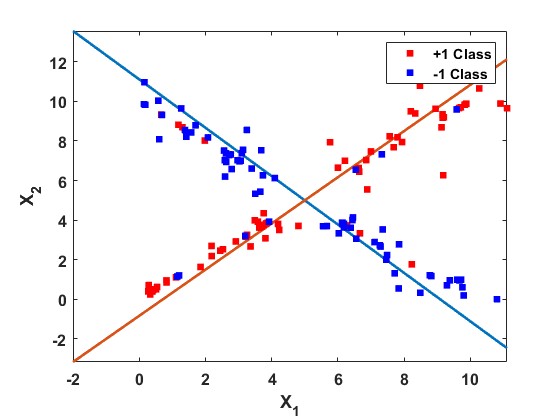}}
\end{minipage}
\begin{minipage}{.30\linewidth}
\centering
\subfloat[IF-IGEPSVM]{\label{fig1:1f}\includegraphics[scale=0.25]{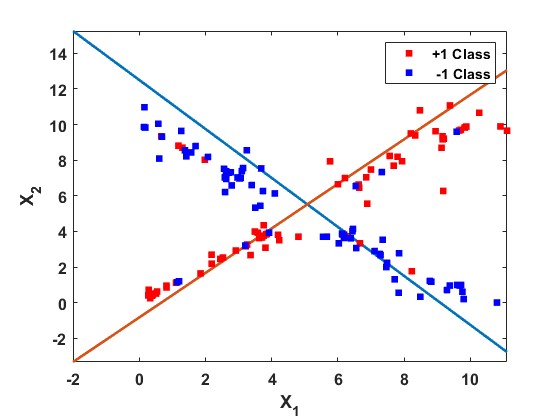}}
\end{minipage}
\caption{The learning outcome on the two-class cross-plane artificial data. Here, $X_1$ and $X_2$ represent the feature of the generated dataset.}
\label{The learning outcome on the artificial data.}
\end{figure*}

We first construct a two-class cross-plane dataset. Lines $X_2=X_1$ and $X_2=-X_1+10$ generate the training dataset for $+1$ and $-1$ classes, respectively. The values of $X_1$ is taken randomly within the range $[0,  4]$ and $[6, 10]$. There are $20$ samples in each of the two classes. We added $8$ outliers to Class $+1$ and $7$ outliers to Class $-1$ to evaluate the robustness. The test dataset is generated in a similar manner to the training data, with the inclusion of uniformly distributed random noise to both the variables $X_1$ and $X_2$. A total of $72$ samples in each of the two classes. We evaluate our proposed IF-GEPSVM and IF-IGEPSVM models along with the baseline models, to this training data. Fig. \ref{The learning outcome on the artificial data.} visually represents the corresponding proximal planes, and Table \ref{classification results of the artificial test data} shows the ACC of the proposed model IF-GEPSVM and IF-IGEPSVM along with the baseline models. The proposed IF-GEPSVM and IF-IGEPSVM models secured the first and second positions with an ACC of $89.28\%$ and $86.04\%$, respectively. As a result, the proposed IF-GEPSVM and IF-IGEPSVM outperformed the baseline models.

Although Fig. \ref{The learning outcome on the artificial data.} looks similar, there are the following major differences that can be observed by a reader: Figs. \ref{fig1:1a}, \ref{fig1:1b}, \ref{fig1:1c}, and \ref{fig1:1d} illustrates that the models do not fit the line across the data points in the cross-place dataset correctly, causing the hyperplane to deviate and misclassify data points. This misalignment is particularly evident in the negative hyperplane, which is far from the negative class data points because some positive class data points are misclassified. These misclassified points are outliers, which deviate significantly from the rest of the data, and adversely affect the model's performance. Therefore, the baseline Pin-GTSVM, GEPSVM, IGEPSVM, and CGFTSVM-ID models tend to yield biased results. However, as shown in Figs. \ref{fig1:1e} and \ref{fig1:1f}, our proposed IF-GEPSVM and IF-IGEPSVM models demonstrate superior classification ability, even after introducing outliers. It utilizes an intuitionistic fuzzy membership scheme to mitigate the impact of noise and outliers, ensuring the hyperplane fits the data points correctly. This robustness to outliers is achieved through a membership and non-membership scheme, which enhances overall model performance. Our proposed models show that they can fit the hyperplane correctly and handle noisy datasets better. These results validate the practicality and feasibility of our IF-GEPSVM and IF-IGEPSVM proposed model, highlighting its improved classification capability in the presence of noise and outliers.

\begin{table}[!ht]
    \centering
    \caption{Classification performance of proposed IF-GEPSVM and IF-IGEPSVM models and baseline models on the artificial dataset.}
    \label{classification results of the artificial test data}
    \resizebox{1\textwidth}{!}{
    \begin{tabular}{lcccccc}
    \hline
\text{Model} $\rightarrow$     & Pin-GTSVM \cite{tanveer2019general}         & GEPSVM \cite{mangasarian2005multisurface}         & IGEPSVM \cite{shao2012improved}  & CGFTSVM-ID \cite{anuradha2024}       & IF-GEPSVM$^{\dagger}$        & IF-IGEPSVM$^{\dagger}$      \\ \hline
Dataset $\downarrow$     & $ACC$ (\%)            & $ACC$(\%)    & $ACC$(\%)        & $ACC$ (\%)              & $ACC$ (\%)              & $ACC$ (\%)              \\
     & $(c_1, c_2)$     & $(\delta)$   & $(\delta, \eta)$  &  $(c_1, c_2, r, \sigma)$    & $(\delta)$     & \textbf{$(\delta, \eta)$}      \\ \hline
Crossplane & 82.27            & \underline{87}             & \underline{87}      &85.78         & \textbf{89.28}            & 86.04            \\
     & $( 2^{-8},  2^{-1})$ & $(2^{-8})$ & $( 2^{-8},  2^{-8})$  & $( 2^{-5},  2^{-1}, 1, 2^{2})$ & $(2^{-8}, 2^{-8})$ & $( 2^{-8},  2^{-8}, 2^{-8})$ \\ \hline
     \multicolumn{7}{l}{$^{\dagger}$ represents the proposed models.}\\
 \multicolumn{7}{l}{Boldface and underline depict the best and second-best models, respectively.}
   \end{tabular}}
\end{table}


\subsection{ Experiments on UCI  and KEEL Datasets}
In this subsection, we present an intricate analysis involving comparison of the proposed IF-GEPSVM and IF-IGEPSVM along with baseline Pin-GTSVM \cite{tanveer2019general}, GEPSVM \cite{mangasarian2005multisurface}, IGEPSVM \cite{shao2012improved} and CGFTSVM-ID models on $62$ benchmark UCI \cite{dua2017uci} and KEEL \cite{derrac2015keel} datasets for the linear and non-linear cases. The classification ACC along with the optimal parameters of the proposed IF-GEPSVM and IF-IGEPSVM models against the baseline models are presented in Table \ref{Average ACC of datasets using linear kernel}. From the table, it is evident that the proposed IF-GEPSVM and IF-IGEPSVM models achieved the highest ACC on the majority of the datasets. The average ACC of the proposed IF-GEPSVM and IF-IGEPSVM models along with the baseline Pin-GTSVM, GEPSVM, IGEPSVM, and CGFTSVM-ID models are $78.20\%$ $81.00\%$, $70.44\%$, $72.33\%$, $74.02\%$ and $77.92\%$, respectively. In terms of average ACC, the proposed IF-IGEPSVM secured the top position, while the proposed IF-GEPSVM achieved the second top position. This observation strongly emphasizes the significant superiority of the proposed models compared to baseline models. As the average ACC can be influenced by exceptional performance in one dataset that compensates for losses across multiple datasets, it might be a biased measure. To mitigate this concern, it becomes essential to individually rank each model for each dataset, enabling a comprehensive assessment of their respective capabilities. In the ranking scheme \cite{demvsar2006statistical}, the model with the poorest performance on a dataset receives a higher rank, whereas the model achieving the best performance is assigned a lower rank.   
Assume that there are $q$ models being evaluated on a total of $N$ datasets. $r_j^i$ represents the rank of the $j^{th}$ model on the $i^{th}$ dataset. Then the average rank of the $j^{th}$ model is calculated as: $R_j = \frac{1}{N} \sum_{i=1}^{N} r_j^i$. The average ranks of the proposed IF-GEPSVM and IF-IGEPSVM models along with the baseline Pin-GTSVM, GEPSVM, IGEPSVM, and CGFTSVM-ID are $2.72$, $2.28$, $4.77$, $4.35$, $3.88$ and $3.01$, respectively. The proposed IF-GEPSVM and IF-IGEPSVM models achieved the lowest average range among the baseline models. Hence the proposed IF-GEPSVM and IF-IGEPSVM models emerged as the better generalization performance. Now we conduct the statistical tests to determine the significance of the results. Firstly, we employ the Friedman test \cite{friedman1940comparison} to determine whether the models have significant differences. Under the null hypothesis, it is presumed that all the models exhibit an equal average rank, signifying equal performance. Friedman statistics follow the chi-square ($\chi_{F}^2$) distribution with $(q-1)$ degree of freedom (d.o.f). The value of $\chi_{F}^2$ is calculated as: $\chi_{F}^2 = \frac{12N}{q(q+1)}\left[ \sum_{i=1}^{q} R_{j}^2 - \frac{q(q+1)^2}{4}\right]$. The $F_F$ statistic follows an $F$-distribution \cite{iman1980approximations} with d.o.f $(q - 1)$ and $(N - 1)(q - 1)$, and is  calculated as: $F_F = \frac{(N-1) \chi_{F}^2}{N(q-1) - \chi_{F}^2}$. For $N=62$ and $q=6$, we get $\chi_{F}^2 = 86.56$ and $F_{F} = 23.63$. From the $F$-distribution table, $F_{F}(5,305) = 2.2435$ at $5\%$ level of significance. Since $F_{F}= 23.63 > 2.2435$, thus we reject the null hypothesis. As a result, there exists a statistical distinction among the models being compared. Now, we employ the Nemenyi post hoc test \cite{demvsar2006statistical} to examine the pairwise distinctions between the models. The value of the critical difference $(C.D.)$ is evaluated as: $C.D. = q_{\alpha}\sqrt{\frac{q(q+1)}{6N}}$, where $q_\alpha$ is the critical value, $C.D.$ is critical difference for $q$ models using $N$ datasets. We get $C.D. = 0.9576$ at $5\%$ level of significance. The average differences in ranking between the proposed IF-GEPSVM and IF-IGEPSVM models along with the baseline Pin-GTSVM, GEPSVM, IGEPSVM, and CGFTSVM-ID models are as follows: $(2.05, 2.49)$, $(1.63, 2.07)$, $(1.16, 1.60)$, and $(0.29, 0.73)$, respectively. As per the Nemenyi post hoc test, the proposed IF-GEPSVM and IF-IGEPSVM models show significant differences compared to the baseline models, except for CGFTSVM-ID. However, the proposed IF-GEPSVM and IF-IGEPSVM models outperform the CGFTSVM-ID model in terms of average rank. Taking into account all these findings, we can conclude that the proposed IF-GEPSVM and IF-IGEPSVM models demonstrate competitive performance compared to the baseline models.

\begin{table}[htp]
    \centering
    \caption{Classification performance of proposed IF-GEPSVM and IF-IGEPSVM models and baseline models on UCI and KEEL datasets with linear kernel.}
    \label{Average ACC of datasets using linear kernel}
    \resizebox{1.0\textwidth}{!}{
    \begin{tabular}{lcccccc}
    \hline
 \text{Model} $\rightarrow$ & Pin-GTSVM \cite{tanveer2019general}         & GEPSVM \cite{mangasarian2005multisurface}         & IGEPSVM \cite{shao2012improved}  & CGFTSVM-ID \cite{anuradha2024}         & IF-GEPSVM$^{\dagger}$       & IF-IGEPSVM$^{\dagger}$       \\  \hline
Dataset $\downarrow$     & $ACC$ (\%)             & $ACC$ (\%)            & $ACC$ (\%)              & $ACC$ (\%)   & $ACC$ (\%)         & $ACC$ (\%)              \\
     & $(c_1, c_2)$     & $(\delta)$   & $(\delta, \eta)$ &  $(c_1, c_2, r, \sigma)$   & $(\delta, \sigma)$     & $(\delta, \eta, \sigma)$     \\ \hline
acute\_nephritis                                      & 79.86        & 78.89       & 74.44   & \textbf{100}     & 80.56       & \underline{88.89}      \\
$(120 \times 6)$                                                     & $(2^2, 2^{-2})$   &  $(2)$           &  $(2^{-4}, 2^{-8})$  &  $(2^{-5}, 2^{-5}, 0.5, 2^{3})$     &   $(2^{-4}, 2^{-8})$          &  $(2^{-7}, 2^{-8}, 2^{-8})$           \\
adult                                                 & 69.78        & 79.63       & 79.72  & 75.56    & \underline{80.87}       & \textbf{89.58}      \\
$(48842 \times 14)$ & $(2, 2^{-4})$     &   $(2^{3})$          &  $(2^{-3}, 2^{-8})$     &  $(2^{-8}, 2^{-6}, 0.5, 2^{4})$       &  $(2^{8}, 2^{-7})$            &  $(2^{-1}, 2^{-8}, 2^{-5})$          \\
aus                          & 78.28        & 88.41       & 88.41  &  \textbf{89.66}    & \underline{88.89}       & 88.41      \\
$(690 \times 15)$                    & $(2^4, 2^6)$   &  $(2^{8})$           &  $(2^{-1}, 2^{-8})$    &  $(2^{2}, 2^{3}, 1, 2^{2})$        & $(2, 2^{-8})$             &     $(2^{-4}, 2^{-8}, 2^{-8})$       \\
bank                                             & 69.03        & 84.37       & 87.61  & 78.66    & \textbf{89.06}       & \underline{88.5}       \\
$(4521 \times 16)$        & $(2, 2)$       &   $(2^{-8})$          &  $(2^{-5}, 2^{-8})$    &  $(2^{-5}, 2^{5}, 0.5, 2^{1})$        & $(2^{8}, 2^{-5})$             &   $(2^{-6}, 2^{-2})$         \\
blood                                                 & 53.88        & 65.63       & \underline{77.23}  & 65.63      & 65.8        & \textbf{78.57}      \\
$(748 \times 4)$           & $(2^{-1}, 1)$    &  $(2^{3})$           &  $(2^{-4}, 2^{-8})$  &  $(2^{3}, 2^{-1}, 1, 2^{-2})$          & $(2^{7}, 2^{-8})$             &   $(2^{-3}, 2^{-8}, 2^{-6})$         \\
breast\_cancer\_wisc\_diag                            & 93.25        & 85.88       & 85.29   & 93.25     & \textbf{95.29}       & \underline{94.71}      \\
$(569 \times 30)$        & $(2^{-1}, 1)$    &   $(2^{5})$          &  $(2^{-3}, 2^{-8})$    &  $(2^{2}, 2^{3}, 0.5, 2^{2})$        & $(2^{2}, 2^{-3})$             &   $(2^{-4}, 2^{-8}, 2^{-6})$         \\
breast\_cancer\_wisc\_prog                            & \underline{81}           & 67.8        & 76.44  & 75.12      & 79.66       & \textbf{86.44}      \\
$(198 \times 33)$         & $(2^{-5}, 2^3)$  &  $(2^{6})$           &  $(2^{-5}, 2^{-8})$   &  $(2^{-1}, 1, 0.5, 2^{-5})$         & $(2^{-3}, 2^{-4})$             &  $(2^{-4}, 2^{4}, 2^{-1})$          \\                                    
breast\_cancer   & 49.75        & 64.71       & 65.88  & \underline{70.83}      & 68.24       & \textbf{84.12}      \\
$(286 \times 9)$        & $(2^6, 2^3)$   &   $(2^{-8})$          &   $(2^{8}, 2^{-8})$   &  $(1, 2^{4}, 0.75, 2^{2})$        &  $(2^{-8}, 2^{-3})$            &   $(2^{-3}, 2^{-2})$         \\
breast\_cancer\_wisc                                  & 97.67        & 87.61       & \underline{98.09}   & \textbf{98.86}     & \underline{98.09}       & \underline{98.09}      \\
 $(699 \times 9)$       & $(1, 2^{-1})$    &   $(2^{8})$          &  $(2, 2^{-6})$   &  $(2^{5}, 2^{-5}, 1, 2^{1})$         &   $(2^{7}, 2^{-8})$           &  $(2^{-2}, 2^{-8}, 2^{-8})$          \\
brwisconsin                                           & 88.78        & 96.08       & 96.57   & \textbf{98.13}     & \underline{97.06}       & \underline{97.06}      \\
$(683 \times 10)$      & $(2^4, 1)$     &     $(2^{7})$        &  $(1, 2^{-8})$  &  $(2^{2}, 2^{-5}, 0.625, 2^{2})$          &   $(2^{7}, 2^{-5})$           &  $(2^{-2}, 2^{7}, 2^{-8}, 2^{-8})$          \\
bupa or liver-disorders                               & 58.56        & 59.22       & 62.14  & \underline{67.99}      & \textbf{78.93}       & \textbf{78.93}      \\  
$(345 \times 7)$           & $(2^{-3}, 2^2) $ &   $(2^{6})$          &  $(2^{-8}, 2^{-8})$  &  $(2^{-5}, 2^{4}, 0.625, 1)$          & $(2^{-8}, 2^{-8})$             & $(2^{-4}, 2^{-8})$           \\
checkerboard\_Data                                    & 87.89        & 88.41       & 88.41  & \textbf{89.66}      & \underline{88.89}       & 88.41      \\
$(690 \times 15)$       & $(2^{-6}, 2^{-4})$ &    $(2^{8})$         & $(2^{-1}, 2^{-8})$     &  $(2^{2}, 2^{3}, 1, 2^{2})$        &  $(2, 2^{-8})$            &    $(2^{-4}, 2^{-8}, 2^{-8})$        \\
chess\_krvkp                                          & \textbf{91.31}        & 66.91       & 71.09  & \underline{86.12}      & 76.72       & 74.43      \\
$(3196 \times 36)$     & $(2^5, 2^{-1})$  &     $(2^{-2})$        &  $(2^{-2}, 2^{-8})$    &  $(2^{4}, 2^{-5}, 0.625, 2^{-5})$        &  $(2^{3}, 2^{-3})$            & $(2^{-2}, 2^{6}, 2^{-8})$           \\
cmc                                                   & 66.19        & 63.95       & 65.99   & 67.06     & \underline{69.61}       & \textbf{71.95}      \\
$(1473 \times 10)$      & $(1, 1)$       &      $(2^{-8})$       &   $(2^{-3}, 2^{-8})$    &  $(1, 2^{4}, 0.5, 2^{2})$       &  $(2^{-8}, 2^{-8})$            &  $(2^{-8}, 2^{-8}, 2^{-8})$          \\
conn\_bench\_sonar\_mines\_rocks                      & 77.42        & 64.52       & 62.9  & 59.68       & \textbf{79.35}       & \underline{78.39}      \\
$(208 \times 60)$    & $(2^2,  2^{-4})$ &     $(2^{-1})$        &  $(2^{2}, 2^{-8})$   &  $(2^{5}, 2^{-5}, 0.5, 2^{4})$         &   $(2^{6}, 2^{-4})$           &   $(2^{6}, 2^{-8}, 2^{-8}, 2^{-8})$         \\                                         
congressional\_voting                                 & 58.63        & 53.08       & 52.31  & 63.27    & \underline{71.54}       & \textbf{75.38}      \\
$(435 \times 16)$      & $(2^{-2}, 2^{-7})$ &   $(2^{6})$          & $(2^{-2}, 2^{-8})$    &  $(2^{1}, 2^{-5}, 0.875, 2^{2})$        &  $(2^{7}, 2^{-8})$            &  $(2^{5}, 2^{-2}, 2^{-8})$          \\
credit\_approval                                      & 74.57        & 81.16       & 83.57 & \textbf{ 87.96}       & \underline{85.99}       & 85.51      \\
$(690 \times 15)$   & $(2^{-1}, 2^3)$  &     $(2^{-8})$        & $(2^{-3}, 2^{-8})$  &  $(2^{-5}, 2^{-4}, 1, 2^{-7})$           &  $(2^{5}, 2^{-8})$            &   $(2^{-4}, 2^{5}, 2^{-8})$         \\
crossplane130                                         & \textbf{100}          & 86.82       & \underline{97.89}  & \textbf{100}      & \textbf{100}         & \textbf{100}        \\
$(130 \times 3)$     & $(2^{-1}, 2^{-4})$ &     $(2^{-8})$        &  $(2^{-8}, 2^{-8})$   &  $(2^{-5}, 2^{-5}, 0.5, 2^{-5})$         &  $(2^{-8}, 2^{-7})$            &   $(2^{-8}, 2^{-8}, 2^{-5})$         \\
crossplane150                                         & 78.26        & 95.72       & \underline{96.78}  &\textbf{100}      & \textbf{100}         & \textbf{100}        \\
$(150 \times 3)$      & $(2^2, 2^{-8})$  &      $(2^{-8})$       &  $(2^{-8}, 2^{-8})$   &  $(2^{-5}, 2^{-5}, 0.5, 2^{-2})$         &   $(2^{-8}, 2^{-3})$           &   $(2^{-8}, 2^{-8}, 2^{-4})$         \\
cylinder\_bands                                       & 50.65        & 66.67       & 67.32  & \underline{67.78}      & 47.06       & \textbf{73.4}       \\
$(512 \times 35)$    & $(2^{-2}, 2^{-8})$ &      $(2^{-8})$       & $(2^{2}, 2^{5})$      &  $(2^{-3}, 2^{2}, 0.875, 2^{-5})$       &  $(2, 2^{-3})$            & $(2^{7}, 2^{-8}, 2^{-5})$           \\
echocardiogram                                        & \underline{90}           & 89.74       & 89.74  & 81.21      & \textbf{94.87}       & \textbf{94.87}      \\
 $(131 \times 10)$       & $(2^2, 2^7)$   &         $(2^{7})$    &  $(2^{-4}, 2^{-8})$     &  $(2^{5}, 2^{-1}, 0.875, 1)$       &   $(1, 2^{-8})$           &   $(2^{-4}, 2^{-8}, 2^{-8})$         \\
fertility                                             & \underline{74.07}        & 50          & 46.67   & 66.67     & \textbf{80}          & \textbf{80}         \\
$(100 \times 9)$       & $(2^{-4}, 2^8)$  &       $(2^{-8})$      & $(2^{6}, 2^{-8})$     &  $(2^{-1}, 2^{-2}, 1, 2^{2})$        &  $(2^{8}, 2^{-3})$            & $(2^{6}, 2^{-8}, 2^{-8}, 2^{-7})$           \\
haber                                                 & 59.72        & 63.74       & 65.93  & 64.32      & \underline{83.63}       & \textbf{84.73}      \\
$(306 \times 4)$       & $(2^{-5}, 2^{-2})$ &      $(2^{7})$       &  $(2^{-2}, 2^{-8})$   &  $(2^{-4}, 2^{3}, 0.75, 2^{3})$         &  $(2^{2}, 2^{-8})$            &  $(2^{-8}, 2^{-8}, 2^{-4})$          \\
haberman\_survival                                    & 59.72        & 63.74       & 65.93  &64.32      & \underline{73.63}       & \textbf{74.73}      \\
$(306 \times 3)$   & $(2^{-5}, 2^{-2})$ &       $(2^{-5})$      & $(2^{-2}, 2^{-8})$    &  $(2^{-4}, 2^{3}, 0.75, 2^{3})$         &   $(2^{2}, 2^{-8}, 2^{-8})$           &  $(2^{-8}, 2^{-8}, 2^{-8})$          \\
heart\_hungarian                                      & 73.69        & 80.68       & \underline{84.09}   & 80.83     & \textbf{86.36}       & \textbf{86.36}      \\
$(294 \times 12)$   & $(2^3, 2^2)$   &     $(2^{3})$        &  $(2^{-1}, 2^{-8})$    &  $(2^{5}, 2^{-1}, 0.875, 2^{2})$        &  $(2, 2^{-4})$            &  $(2, 2^{4}, 2^{-3})$          \\ 
horse\_colic                                          & 74.84        & 74.55       & 76.36  & \textbf{82.93}      & 80.91       & \underline{82.73}      \\
$(368 \times 25)$  & $(2^{-2}, 1)$    &    $(2^{8})$         &   $(1, 2^{-8})$    &  $(2^{3}, 2^{-2}, 1, 2^{3})$       &  $(2^{4}, 2^{-5})$            &   $(2^{2}, 2^{4}, 2^{-8})$         \\
ilpd\_indian\_liver                                   & 60.54        & 47.7        & \underline{64.37}   &63.85     & 55.75       & \textbf{72.41}      \\
$(583 \times 9)$  & $(2^{-6}, 2)$    &     $(1)$        &   $(2^{-1}, 2^{-8})$   &  $(2^{2}, 2^{2}, 0.5, 2^{8})$        & $(2^{6}, 2^{-8})$             &    $(1, 2^{-6}, 2^{-8})$        \\
hepatitis                                             & 63.74        & 69.57       & 76.09   &67.58     & \underline{82.61}       & \textbf{84.78}      \\
$(155 \times 19)$      & $(2^{-1}, 2^{-6})$ &      $(2^{6})$       & $(2^{-4}, 2^{-8})$    &  $(2^{-3}, 2^{-1}, 0.625, 2^{2})$         &  $(2^{4}, 2^{-6})$            & $(2^{-1}, 2^{-8}, 2^{-7})$           \\ 
hill\_valley                              & 54.78        & 59.23       & 56.2    &\textbf{88.43}     & \underline{60.14}       & 58.73      \\
$(1212 \times 100)$    & $(2, 2^7)$     &      $(1)$       &  $(2^{-6}, 2^{-8})$    &  $(2^{-3}, 2^{3}, 0.625, 2^{2})$        &  $(2^{-7}, 2^{-8})$            &   $(2^{-8}, 2^{-8}, 2^{-8})$         \\ 
iono                                                  & \textbf{85.99}        & 80          & 74.29   & \underline{85.27}     & 78.1        & 73.33      \\
$(351 \times 34)$    & $(2^{-8}, 2^{-1})$ &     $(2^{2})$        &  $(2, 2^{-8})$   &  $(2^{-4}, 2^{-3}, 0.75, 2^{5})$         &  $(2^{-1}, 2^{-8})$            & $(2^{4}, 2^{-8}, 2^{-8})$           \\
ionosphere                                            & 85.46        & 81.9        & 74.29  & \underline{86.78}      & 78.1        & \textbf{90}         \\
$(351 \times 33)$   &$ (2^{-2}, 2^2)$  &     $(2)$        & $(2^{-4}, 2^{-8})$    &  $(2^{-4}, 2^{-5}, 1, 2^{2})$         &  $(2, 2^{-8})$            &   $(2^{7}, 2^{-8}, 2^{-5})$         \\ \hline
\multicolumn{7}{l}{$^{\dagger}$ represents the proposed models.}\\
 \multicolumn{7}{l}{Boldface and underline depict the best and second-best models, respectively.}
 \end{tabular}}
\end{table}

\begin{table}[htp]
    \centering
    \caption*{\text{Table 2: } (Continued)}
    \resizebox{1.0\textwidth}{!}{
    \begin{tabular}{lcccccc}
    \hline
\text{Model} $\rightarrow$ & Pin-GTSVM \cite{tanveer2019general}         & GEPSVM \cite{mangasarian2005multisurface}         & IGEPSVM \cite{shao2012improved}  & CGFTSVM-ID \cite{anuradha2024}         & IF-GEPSVM$^{\dagger}$        & IF-IGEPSVM$^{\dagger}$       \\  \hline
Dataset $\downarrow$     & $ACC$ (\%)              & $ACC$ (\%)            & $ACC$ (\%)      & $ACC$ (\%)        & $ACC$ (\%)              & $ACC$ (\%)              \\
     & $(c_1, c_2)$     & $(\delta)$   & $(\delta, \eta)$  &  $(c_1, c_2, r, \sigma)$    & $(\delta, \sigma)$    & $(\delta, \eta, \sigma)$     \\ \hline
magic                                                 & 65.78        & 74.71       & 74.17  & \underline{74.89}      & \textbf{76.31}       & 72.51      \\
$(19020 \times 10)$  & $(2^{-4}, 2^3)$  &       $(2^{5})$      &  $(2^{-2}, 2^{-8})$  &  $(2^{-8}, 2^{-8}, 1, 2^{4})$          & $(2^{8}, 2^{-8})$             &  $(2^{-1}, 2^{-8}, 2^{-8})$          \\
monks\_1                                              & 69.27        & \underline{76.75}       & 56.75 &72.51       & 66.27       & \textbf{84.34}      \\
$(556 \times 6)$    & $(2^{-2}, 2^{-1})$ &    $(2^{-8})$         & $(2^{4}, 2^{-8})$   &  $(1, 2^{-1}, 1, 2^{2})$          &  $(2^{-8}, 2^{-8})$            &  $(2^{-6}, 2^{-8}, 2^{-8})$          \\
monks\_2                                              & 50           & 56.11       & \textbf{57.22}  &55.66      & 56.11       & \underline{56.67}      \\
$(601 \times 6)$   & $(2^{-3}, 2^{-1})$ &   $(2^{6})$          & $(2^{-5}, 2^{-8})$     &  $(2^{5}, 2^{2}, 0.625, 2^{5})$        & $(2^{6}, 2^{-4})$             &   $(2^{-8}, 2^{-8}, 2^{-4})$         \\
monks\_3                                              & 79.75        & \textbf{82.53}       & 80.12   &80.21     & 71.08       & \underline{80.72}      \\
$(554 \times 6)$  & $(2, 2^6)$     &      $(2^{-8})$       &  $(2^{-5}, 2)$      &  $(2^{5}, 2^{2}, 0.875, 2^{-5})$      &   $(2^{4}, 2^{-3})$           &   $(2, 2^{-8}, 2^{-8})$         \\                                                      
mammographic                                          & 64.02        & 70.49       & 81.6  & \textbf{83.16}       & \underline{82.99}       & 80.9       \\
$(961 \times 5)$   & $(2^{-8}, 1)$    &     $(2^{5})$        &  $(2^{4}, 2^{7})$   &  $(2^{5}, 2^{5}, 0.5, 2^{-5})$         & $(2^{2}, 2^{-3})$             &  $(2^{-1}, 2^{-8}, 2^{-5})$          \\
molec\_biol\_promoter                                 & 57.48        & 54.84       & 51.61   & \textbf{76.92}     & 41.94       & \underline{71.94}      \\
$(106 \times 57)$ & $(2^{-4}, 2^{-4})$ &      $(2^{-6})$       & $(2^{-1}, 2^{-8})$   &  $(2^{3}, 2^{-5}, 0.5, 2^{5})$          &  $(2^{4}, 2^{-8})$            &  $(2^{-7}, 2^{-8}, 2^{-8})$          \\
mushroom   & 65.45        & \textbf{97.62}       & 96.8   &95.85      & \underline{97.50}        & 92.89      \\
$(8124 \times 21)$  & $(2^{-4}, 2^{-2})$  &    $(2)$         & $(2^{-8}, 2^{-8})$   &  $(2^{-5}, 2^{-5}, 0.5, 2^{7})$          &  $(2^{-8}, 2^{-8})$            &   $(2^{-5}, 2^{-8}, 2^{-8})$         \\
oocytes\_merluccius\_nucleus\_4d                      & 68.3        & 70.92       & 68.3    & \textbf{82.84}     & 71.9        & \underline{79.67}       \\
$(1022 \times 41)$    & $(2, 2^{-3})$    &    $(2^{7})$         & $(2^{-2}, 2^{-8})$    &  $(2^{-4}, 2^{1}, 0.5, 2^{5})$         &  $(2^{-7}, 2^{-7})$            &    $(2^{-2}, 2^{-8}, 2^{-7})$        \\
oocytes\_trisopterus\_nucleus\_2f                     & 72.76        & 57.51       & 60.07  & \underline{79.06}       & 67.77       & \textbf{88.97}      \\
$(912 \times 25)$   & $(2^7, 2^{-5})$  &    $(2^{-1})$         &  $(2^{-1}, 2^{-8})$     &  $(2^{-3}, 2^{2}, 1, 2^{4})$       &  $(2^{5}, 2^{-8})$            &  $(2^{-1}, 2^{3}, 2^{-8})$          \\                                                
musk\_1      & 76.19        & 66.2        & 63.38    & \underline{78.94}    & \textbf{83.8}        & 54.23      \\
$(476 \times 166)$     & $(2^{-4}, 2^5)$  &     $(2^{8})$        & $(2^{-3}, 2^{4})$   &  $(1, 2^{-1}, 1, 2^{2})$          &  $(2^{3}, 2^{-8})$            &  $(2^{7}, 2^{-3}, 2^{-4})$         \\
musk\_2                                               & 81.65        & 87.01       & 84.03   &80.53     & \underline{94.78}       & \textbf{95.46}      \\
$(6598 \times 166)$  & $(2^{-8}, 2^{-7})$ &   $(2^{8})$          & $(2^{-4}, 2^{-8})$   &  $(2^{5}, 2^{2}, 0.625, 2^{5})$          &  $(2^{6}, 2^{-7})$            &   $(2^{2}, 2^{-2}, 2^{-8})$         \\
ozone                                                 & 80.78        & 94.61       & \textbf{97.24}  &78.41      & 84.56       & \underline{96.85}      \\
$(2536 \times 72)$  & $(2^{-1}, 2^{-4})$ &   $(2^{2})$          &  $(2^{-5}, 2^{-8})$     &  $(2^{-5}, 2^{4}, 0.5, 2^{4})$       & $(2^{-3}, 2^{-5})$             &  $(2^{-2}, 2^{-8}, 2^{-4})$          \\
parkinsons                                            & 69.15        & 60.34       & 77.59  & \textbf{86.75}      & \underline{82.76}       & 77.59      \\
$(195 \times 22)$    & $(2, 2^{-4})$    &     $(2^{6})$        & $(2^{-8}, 2^{-8})$     &  $(2^{-3}, 2^{-5}, 0.625, 2^{4})$        &  $(2^{-3}, 2^{-8})$            &  $(2^{-8}, 2^{-8}, 2^{-7})$          \\
pima                                                  & 50           & 76.09       & \underline{77.83}  &73.21      & 76.09       & \textbf{83.91}      \\
$(768 \times 8)$    & $(2^2, 2^{-4})$  &   $(2^{5})$          &  $(2^{-4}, 2^{2})$     &  $(2^{-5}, 2^{-2}, 0.25, 2^{-4})$       &   $(2^{3}, 2^{-4})$           & $(2^{-3}, 2^{-8}, 2^{-7})$           \\
pittsburg\_bridges\_T\_OR\_D                          & 43.33        & \underline{50}          & \underline{50}  & 45         & \textbf{63.33}       & \underline{50}         \\
$(102 \times 7)$  & $(2^2, 2^{-4})$  &   $(2^{7})$          &  $(2, 2^{-8})$   &  $(1, 2^{-2}, 1, 2^{2})$         &    $(1, 2^{-8})$          &    $(2^{3}, 2^{-8}, 2^{-4})$        \\
planning                                              & 44.44        & \underline{64.81}       & \textbf{68.52}  & 40.28      & 55.56       & 61.11      \\
$(182 \times 12)$   & $(2^{-8}, 2^2)$  &   $(2^{-7})$          &  $(2^{-3}, 2^{-8})$    &  $(2^{-5}, 2^{-5}, 0.875, 2^{3})$        &   $(2^{-4}, 2^{-3})$           &  $(2^{-4}, 2^{6}, 2^{-8})$          \\
ringnorm                                              & 62.76        & \underline{75.99}       & \textbf{76.8 }   &75.28     & 75.85       & 75.97      \\
$(7400 \times 20)$    & $(2^{-2}, 2^3)$  &    $(2^{8})$         &  $(2^{6}, 2^{-8})$   &  $(2^{-8}, 2^{-7}, 1, 2^{7})$         &  $(2^{7}, 2^{-4})$            &   $(1, 2^{-8}, 2^{-1})$         \\
sonar          & 66.19        & 62.9        & 66.13   & \underline{73.76}     & \textbf{74.19}       & \textbf{74.19}      \\
 $(208 \times 61)$   & $(2^{-7}, 1)$    &   $(2^{5})$          &  $(2^{-7}, 2^{-8})$     &  $(2^{5}, 2^{3}, 0.875, 2^{-5})$       &  $(2^{2}, 2^{-8})$            &   $(2^{-4}, 2^{-8}, 2^{-5})$         \\ 
spambase                                              & 61.79        & 64.06       & 63.48  & 81.45      & \textbf{87.86}       & \underline{83.11}      \\
$(4601 \times 57)$    & $(2^{-3}, 2^4)$  &  $(2^{-8})$           & $(2^{-6}, 2^{-8})$     &  $(2^{-8}, 2^{-8}, 0.5, 2^{8})$        &   $(2^{-2}, 2^{-8})$           &   $(2^{-3}, 2^{-8}, 2^{-8})$         \\   
statlog\_australian\_credit                           & 50.79        & 55.07       & 62.8    & \underline{58.63}     & 55.07       & \textbf{64.25}      \\
$(690 \times 14)$   & $(2^3, 2^{-1})$  &  $(2^{8})$           & $(1, 2^{-8})$   &  $(2^{5}, 2^{-2}, 1, 2^{2})$          &   $(2^{7}, 2^{-4})$           &  $(2^{-1}, 2^{-3}, 2^{-3})$          \\
statlog\_german\_credit                               & 68.51        & 72          & 70.33   &77.42     & \textbf{80}          & \underline{79.67}      \\
$(1000 \times 24)$     & $(1, 2^{-4})$    &   $(2^{3})$          & $(2^{-2}, 2^{-8})$     &  $(2^{2}, 2^{1}, 1, 2^{1})$        &  $(2^{5}, 2^{-3})$            &  $(2^{-3}, 2^{-8}, 2^{-3})$          \\
spect                                                 & 59.84        & 53.16       & 55.7   & \textbf{71.57}      & \underline{66.96}       & 65.7       \\
$(265 \times 22)$   & $(2^{-1}, 2^7)$  &   $(2^{-1})$          &  $(2^{-4}, 2^{-8})$  &  $(2^{4}, 1, 1, 2^{3})$          & $(2^{-4}, 2^{-3})$             &  $(2^{-8}, 2^{-8}, 2^{-8})$          \\
spectf                                                & \underline{81.09}        & 66.25       & 75     & \textbf{85.06}      & 72.5        & 80         \\
$(267 \times 44)$    & $(2, 2^8)$     &   $(1)$          &  $(2^{-5}, 2^{6})$   &  $(2^{1}, 2^{-3}, 0.75, 2^{4})$         &   $(2^{-5}, 2^{-8})$           &  $(2^{-5}, 2^{7}, 2^{-4})$          \\
statlog\_heart                                        & 81.44        & \underline{83.95}       & \underline{83.95}  & \textbf{86.08}      & 81.48       & 81.48      \\
$(270 \times 13)$  & $(2^{-3}, 2^{-3})$ &  $(1)$           &  $(2^{-1}, 2^{-8})$   &  $(2^{4}, 2^{-1}, 0.75, 2^{-5})$         &   $(2^{2}, 2^{-6})$           &    $(2^{-8}, 2^{-8}, 2^{-8})$        \\
titanic                                               & 50           & \textbf{78.03}       & \textbf{78.03}   &70.51     & \underline{77.27}       & 74.24      \\
$(2201 \times 3)$ & $(2^{-5}, 2^{-8})$ &   $(2^{7})$          &  $(2^{-2}, 2^{-8})$   &  $(2^{-5}, 2^{-3}, 0.625, 2^{2})$         &   $(2^{6}, 2^{-4})$           &   $(2^{2}, 2^{-8}, 2^{-3})$         \\
twonorm                                               & 82.87        & 87.79       & 87.7   &85.75      & \underline{97.9}        & \textbf{98.05}      \\
$(7400 \times 20)$  & $(2^{-8}, 2^{-3})$ &  $(2^{7})$           &  $(2^{-7}, 2^{4})$ &  $(2^{-8}, 2^{-8}, 0.5, 2^{6})$           &   $(2^{6}, 2^{-8})$           &   $(1, 2^{-8}, 2^{-8})$         \\   
tic\_tac\_toe                                         & 76.81        & 71.43       & 73.52     & \underline{96.81}   & \textbf{97.91}       & \textbf{97.91}      \\
$(958 \times 9)$      & $(2^3, 2)$     &     $(2^{-8})$        & $(2^{-8}, 2^{-8})$     &  $(2^{-5}, 2^{-5}, 0.5, 2^{-5})$        &   $(2^{-8}, 2^{-8})$           &   $(2^{-6}, 2^{4}, 2^{-3})$         \\
vehicle1                                              & 70.13        & 75.49       & 76.28   & \textbf{81.18}     & \underline{77.87}       & \underline{77.87}      \\
$(846 \times 19)$     & $(2^5, 2^7)$   &   $(2^{-6})$          &  $(2^{-6}, 2^{-8})$   &  $(2^{-1}, 1, 0.5, 2^{1})$         &  $(2^{-1}, 2^{-4})$            &   $(1, 2^{-8}, 2^{-8})$         \\
vertebral\_column\_2clases                            & 68.82        & 70.97       & 67.74   & \underline{76.27}     & \textbf{76.34}       & 73.41      \\
$(310 \times 6)$   & $(2^{-8}, 2^{-8})$ &   $(2^{-8})$          &  $(2^{2}, 2^{-8})$    &  $(2^{-1}, 2^{5}, 0.75, 2^{3})$        &   $(2^{4}, 2^{-3})$           &  $(2^{3}, 2^{7}, 2^{-4})$          \\ 
vowel         & 90.87        & 78.18       & 80.2   & \underline{91.06}      & 89.86       & \textbf{92.57}      \\
$(988 \times 11)$    & $(2^{-8}, 2^4)$  &   $(2^{2})$          &   $(2^{-7}, 2^{-8})$      &  $(2^{5}, 2^{1}, 0.75, 2^{-5})$     &   $(2^{-3}, 2^{-4})$           &  $(2^{-8}, 2^{-8}, 2^{-8})$          \\
wpbc      & \underline{69.69}        & 62.07       & 74.14  &68.53      & 68.97       & \textbf{72.41}      \\
$(194 \times 34)$  & $(2^{-8}, 2^4)$  &   $(2^{-6})$          &  $(2^{-5}, 2^{-8})$  &  $(2^{1}, 1, 0.625, 2^{-5})$          &   $(2^{-3}, 2^{-6})$           &    $(2^{-8}, 2^{-8}, 2^{-8})$        \\ \hline
Average $ACC$ & 70.44  & 72.33 & 74.02 & 77.92  & \underline{78.20} & \textbf{81.00} \\ \hline
Average Rank & 4.77  & 4.35 & 3.88 & 3.01  & \underline{2.72}  & \textbf{2.28}  \\  \hline 
\multicolumn{7}{l}{$^{\dagger}$ represents the proposed models.}\\
 \multicolumn{7}{l}{Boldface and underline depict the best and second-best models, respectively.}
\end{tabular}}
\end{table}

\begin{table}
    \centering
    \caption{Pairwise win-tie-loss test of proposed IF-GEPSVM and IF-IGEPSVM models along with baseline models on UCI and KEEL datasets with linear kernel.}
 \label{Pairwise Win-tie using linear kernel}
    \resizebox{1.0\textwidth}{!}{
    \begin{tabular}{lcccccc}
    \hline
           & Pin-GTSVM \cite{tanveer2019general}       & GEPSVM \cite{mangasarian2005multisurface}          & IGEPSVM \cite{shao2012improved} &CGFTSVM-ID \cite{anuradha2024}       & IF-GEPSVM$^{\dagger}$        \\ \hline
GEPSVM \cite{mangasarian2005multisurface}    & {[}37, 0, 25{]} &                 &                 &       &           \\
IGEPSVM \cite{shao2012improved}   & {[}39, 1, 22{]} & {[}36, 6, 20{]} &                 &        &          \\
CGFTSVM-ID \cite{anuradha2024}   &  {[}50, 2, 10{]}    & {[}44, 1, 17{]}    & {[}39, 0, 23{]}    &    &       \\
IF-GEPSVM$^{\dagger}$  & {[}48, 1, 13{]} & {[}47, 3, 12{]} & {[}46, 1, 15{]} &  {[}33, 2, 27{]}         &          \\
IF-IGEPSVM$^{\dagger}$ & {[}57, 1, 4{]}  & {[}49, 3, 12{]} & {[}45, 5, 12{]} &  {[}38, 2, 22{]}  & {[}30, 12, 20{]}    \\ \hline
\multicolumn{6}{l}{$^{\dagger}$ represents the proposed models.}
\end{tabular}}
\end{table}

\begin{table}[htp]
    \centering
    \caption{Classification performance of proposed IF-GEPSVM and IF-IGEPSVM models and baseline models on UCI and KEEL datasets with non-linear kernel.}
    \label{Average ACC of datasets using Gaussian kernel}
   \resizebox{1.0\textwidth}{!}{
    \begin{tabular}{lcccccc}
    \hline
 \text{Model} $\rightarrow$ & Pin-GTSVM \cite{tanveer2019general}         & GEPSVM \cite{mangasarian2005multisurface}         & IGEPSVM \cite{shao2012improved}   &  CGFTSVM-ID \cite{anuradha2024}  & IF-GEPSVM${^\dagger}$       & IF-IGEPSVM${^\dagger}$      \\  \hline
Dataset $\downarrow$     & $ACC$ (\%)              & $ACC$ (\%)   & $ACC$ (\%)         & $ACC$ (\%)             & $ACC$ (\%)              & $ACC$ (\%)              \\
     & $(c_1, c_2, \sigma)$     & $(\delta, \sigma)$   & $(\delta, \eta, \sigma)$ & $(c_1, c_2, r, \sigma)$    & $(\delta, \sigma)$     & $(\delta, \eta, \sigma)$      \\ \hline
acute\_nephritis                  & 86.11        & 93.78  & 91.64 & \textbf{100}  & \underline{97}     & \textbf{100}        \\
$(120 \times 6)$      & $(2^{-8}, 2^{-8}, 2^{-4})$         & $(2, 2^{-8})$       & $(2^{-8}, 2^{-8}, 2^{-1})$  &  $(2^{-5}, 2^{-5}, 0.5, 2^{-1})$      &  $(2^{-8}, 2^{-2})$         & $(2^{-8}, 2^{-8}, 2^{-2})$           \\
adult      & 76.67    & 78.65  & 75.45 & 80.48   & \underline{84.46}     & \textbf{86.65}      \\
$(48842 \times 14)$         & $(2^{-7}, 2^{-8}, 2^{2})$         & $(2^{4}, 2^{7})$       &  $(2^{-4}, 2^{-5}, 2^{3})$   &  $(2^{-4}, 2^{-1}, 1, 2^{-2})$    & $(2^{-4}, 2^{-7})$          &  $(2^{-7}, 2^{-4}, 2^{-1})$          \\
aus       & 85.02    & 85.99  & \textbf{88.38}  & 85.2   & \underline{87.21}     & \textbf{88.38}      \\
$(690 \times 15)$       & $(2^{-6}, 2^{-5}, 2^{7})$         & $(2^{4}, 2^{3})$       &  $(2^{-3}, 2^{-8}, 2^{7})$ &  $(2^{-3}, 2^{-4}, 0.625, 2^{3})$      & $(2, 2^{2})$          &    $(2^{-3}, 2^{7}, 2^{-1})$        \\
bank         & 72.65    & 77.89  & 88.35 & 70.94  & \textbf{90.81}     & \underline{88.42}      \\
$(4521 \times 16)$     & $(2^{-4}, 2^{3}, 2^{-4})$         & $(2^{-6}, 2^{2})$       & $(2^{-3}, 2^{-8}, 2^{2})$  &  $(2^{-1}, 2^{-5}, 1, 2^{3})$       & $(2^{5}, 2^{-3})$          &  $(1, 2^{-8}, 2^{2})$          \\
blood       & 59.28    & 57.59  & \underline{77.68}  & 53.6  & \textbf{78.57}     & \textbf{78.57}      \\
$(748 \times 4)$   & $(2^{-6}, 2^{-1}, 2)$         & $(2^{2}, 2^{-4})$       & $(2^{-2}, 2^{-5}, 2^{-3})$  &  $(2^{-4}, 2^{2}, 0.625, 2^{2})$      & $(2^{-8}, 2^{4})$          &   $(1, 2^{-8}, 2^{2})$         \\
breast\_cancer         & 54.35    & 68.24  & \textbf{72.94}  & \underline{70.89}  & \textbf{72.94}     & \textbf{72.94}      \\
$(286 \times 9)$      & $(2^{-8}, 2^{-8}, 2^{-8})$    & $(2^{8}, 2^{3})$       &  $(1, 2^{-8}, 2^{-8})$  &  $(2^{-5}, 2^{-5}, 0.5, 2^{-1})$     &  $(2^{-2}, 2^{6})$         &     $(2^{-1}, 2^{-8}, 2^{6})$       \\
breast\_cancer\_wisc\_diag        & \underline{94.43}    & 75.88  & \textbf{94.71}  & 90.08  & 74.12     & 93.53      \\
 $(569 \times 30)$       &  $(2^{-1}, 2^{3}, 2^{6})$        & $(2^{3}, 2^{-3})$       &  $(2^{-6}, 2^{-8}, 2^{7})$  &  $(1, 2^{1}, 1, 2^{3})$     &  $(1, 2^{5})$         &  $(2^{-8}, 2^{-8}, 2^{7})$          \\
breast\_cancer\_wisc\_prog        & 64.22    & 69.49  & \textbf{86.44}  & 64.83  & 71.19     & \underline{83.05}      \\
$(198 \times 33)$      & $(2, 2^{-8}, 2^{7})$         & $(2^{2}, 1)$       &  $(2^{-1}, 2^{-6}, 2^{2})$  &  $(2^{-5}, 2^{3}, 0.875, 2^{3})$     &  $(2^{-8}, 2^{-8})$         &   $(2^{-7}, 2^{-8}, 2)$         \\                                  
breast\_cancer\_wisc              & 95.24    & 88.52  & 88.56 & \underline{98.05}   & 92.6      & \textbf{98.09}      \\
$(699 \times 9)$     & $(2^{-3}, 2^{7}, 2^{3})$         & $(2^{5}, 2^{8})$       & $(2^{4}, 2^{-8}, 2^{3})$  &  $(2^{-5}, 2^{5}, 0.875, 2^{4})$      & $(2^{2}, 2^{2})$         &  $(2^{5}, 2^{-8}, 2^{3})$          \\
brwisconsin                       & \underline{95.96}    & 87.25  & 76.23  & 88.74  & 89.22     & \textbf{96.23}      \\
$(683 \times 10)$     & $(2^{-8}, 1, 2^{-1})$         & $(2^{4}, 2^{2})$       & $(2^{2}, 2^{-8}, 2^{6})$   &  $(2^{-3}, 1, 0.5, 2^{3})$     &  $(2^{-1}, 2^{6})$         &  $(2^{2}, 2^{-8}, 2^{6})$          \\
bupa or liver-disorders           & 66.31    & 67.96  & 60.58 & 66.85  & \underline{68.93}     & \textbf{74.58}      \\
$(345 \times 7)$  & $(2^{-4}, 2^{-4}, 2^{5})$         & $(2^{-4}, 2^{3})$       & $(2^{-3}, 2^{-8}, 2^{-2})$  &  $(2^{-5}, 2^{-1}, 1, 2^{2})$      &  $(2^{-2}, 2^{-1})$         &  $(2^{-3}, 2^{-8}, 2^{-2})$          \\
checkerboard\_Data                & 87.21    & 82.61  & \underline{88.38} & \textbf{90.2}  & 84.06     & \underline{88.38}      \\
$(690 \times 15)$      &  $(2^{-6}, 2^{-5}, 2^{7})$        & $(2^{6}, 1)$       &  $(2^{-3}, 2^{-8}, 2^{7})$  &  $(2^{-3}, 2^{-4}, 0.625, 2^{3})$     &  $(2, 2^{-2})$         &   $(2^{-3}, 2^{-8}, 2^{7})$         \\
chess\_krvkp                      & 66.76    & 65.8   & 66.49 & 70.16  & \underline{70.85}     & \textbf{75.67}      \\
$(3196 \times 36)$ &  $(2^{-3}, 2^{-8}, 2^{-4})$        & $(2^{4}, 2^{-3})$       &  $(2^{-7}, 2^{-8}, 2^{3})$  &  $(2^{-5}, 2^{3}, 0.625, 2^{3})$     &  $(2, 2^{4})$         &   $(2^{-6}, 2^{-8}, 2^{5})$         \\
cmc                               & 67.03    & 63.23  & 57.61 & 63.83  & \underline{69.61}     & \textbf{74.61}      \\
$(1473 \times 10)$     &  $(2^{-2}, 1, 2^{5})$        & $(2^{-2}, 2^{4})$       & $(2^{2}, 2^{-8}, 2^{6})$ &  $(2^{5}, 2^{2}, 0.875, 2^{4})$       & $(2^{-2}, 2)$          &    $(2^{2}, 2^{-8}, 2^{6})$        \\
conn\_bench\_sonar\_mines\_rocks  & 50.65    & 54.84  & 69.35 & 54.84   & \underline{76.45}     & \textbf{80.87}      \\
$(208 \times 60)$  & $(2^{-8}, 1, 2^{8})$         &  $(2^{7}, 2^{-2})$      &   $(2^{3}, 2^{-8}, 2^{8})$  &  $(2^{1}, 2^{1}, 0.5, 2^{5})$    &  $(2^{-5}, 2^{7})$         &   $(2^{3}, 2^{-4}, 2^{3})$         \\                                  
congressional\_voting             & 53.66    & 60     & \textbf{64.62}  & 62.35  & \underline{63.08}     & 57.69      \\
$(435 \times 16)$    &  $(2^{-4}, 2^{-3}, 2^{-2})$        & $(2^{8}, 2^{-1})$       & $(2^{-1}, 2^{-8}, 2^{2})$ &  $(2^{-3}, 2^{-4}, 1, 2^{1})$       &  $(2^{-6}, 2^{6})$         &  $(2, 2^{-8}, 2^{2})$          \\
credit\_approval                  & 62.45    & 51.06  & 63.09 & \textbf{86.55}   & 69.42     & \underline{73.09}      \\
$(690 \times 15)$   & $(2^{4}, 2^{2}, 2^{4})$         & $(2^{7}, 2^{-1})$       & $(2^{5}, 2^{-8}, 2^{2})$ &  $(2^{4}, 1, 0.75, 2^{2})$       &  $(2^{-8}, 2^{8})$         &  $(2^{7}, 2^{-8}, 2^{2})$          \\
crossplane130                     & 92.87    & 91.65  & 93.65 & \underline{95}  & \textbf{95.89}     & \textbf{95.89}      \\
$(130 \times 3)$    & $(2^{-6}, 2^{-3}, 2^{3})$         & $(2^{-5}, 2^{-1})$       &  $(2^{-8}, 2^{-8}, 1)$  &  $(2^{-5}, 2^{-5}, 0.5, 2^{-1})$     & $(1, 1)$          &      $(2^{-8}, 2^{-8}, 1)$      \\
crossplane150                     & 87.78    & 89.68  & 80.45 & 95.73  & \underline{96.67}     & \textbf{99.56}      \\
$(150 \times 3)$  &  $(2^{-1}, 1, 2^{-6})$        &  $(2, 2^{6})$      & $(2^{-8}, 2^{-8}, 1)$     &  $(2^{-5}, 2^{-5}, 0.5, 2^{-3})$   &  $(2^{6}, 2)$         &   $(2^{-8}, 2^{-8}, 1)$         \\
cylinder\_bands                   & 62.3     & 60.78  & 64.05  & \textbf{71.33}  & 65.86     & \underline{66.01}      \\
$(512 \times 35)$  & $(2^{-4}, 2^{-3}, 2^{3})$         & $(2^{3}, 2^{-5})$       & $(2^{-3}, 2^{-8}, 2^{2})$  &  $(2^{-4}, 1, 0.5, 2^{2})$      &  $(2^{-8}, 2^{-1})$         &   $(2^{3}, 2^{-8}, 1)$         \\
echocardiogram                    & 77.93    & 66.67  & \underline{89.74} & 86.38  & 69.23     & \textbf{94.87}      \\
$(131 \times 10)$ &  $(2^{-8}, 2^{2}, 2^{6})$         & $(2^{3}, 2^{-2})$       & $(2^{-2}, 2^{-8}, 2^{3})$  &  $(2^{-5}, 1, 0.5, 2^{5})$      &   $(2^{-3}, 2^{7})$        &    $(2^{-4}, 2^{-8}, 2^{2})$        \\
fertility                         & 60       & 66.67  & 76.83 & 50  & \textbf{90}        & \underline{86.67}      \\
$(100 \times 9)$  & $(2^{-8}, 2^{-8}, 1)$         & $(2^{8}, 2)$       & $(2^{-8}, 2^{-8}, 1)$      &  $(2^{-2}, 2^{-3}, 0.125, 2^{-1})$  &   $(2^{-8}, 2^{-8})$        &    $(2^{-8}, 2^{-8}, 1)$        \\
haber                             & 53.36    & 71.43  & \underline{74.77} & 58.63  & \textbf{78.57}     & \underline{74.77}      \\
$(306 \times 4)$   & $(2^{2}, 2, 2^{8})$         &  $(2^{6}, 2^{-5})$      & $(2^{-3}, 2^{-5}, 2)$ &  $(2^{4}, 2^{2}, 0.75, 2^{1})$       &   $(2^{-3}, 2^{4})$        &    $(2^{-3}, 2^{-8}, 2)$        \\
haberman\_survival                & 53.36    & 70.33  & \underline{75.82} & 58.63   & 70.33     & \textbf{76.92}      \\
$(306 \times 3)$   & $(2^{2}, 2, 2^{8})$         & $(2^{-7}, 2^{6})$       &  $(1, 2^{-8}, 1)$     &  $(2^{4}, 2^{2}, 0.75, 2^{1})$  &  $(2^{5}, 2^{4})$         &     $(2, 2^{-8}, 2)$       \\
heart\_hungarian                  & 81.79    & 77.27  & 84.09 & 83.57  & \textbf{89.77}     & \underline{85.23}      \\
$(294 \times 12)$   & $(2^{-2}, 1, 2^{8})$         &  $(2^{7}, 2^{3})$      &  $(2^{3}, 2^{-8}, 2^{8})$ &  $(2^{5}, 2^{3}, 1, 2^{5})$      &   $(2^{-8}, 2^{8})$        &   $(2^{8}, 2^{-8}, 2^{7})$         \\ 
horse\_colic                      & 76.88    & 83.64  & 82.45 & \textbf{85.75}  & \underline{85.45}     & 84.73      \\
$(368 \times 25)$     & $(2^{-4}, 2^{-1}, 2^{8})$         & $(2^{6}, 2^{3})$       & $(2^{-3}, 2^{-8}, 2^{5})$  &  $(2^{-1}, 2^{4}, 0.625, 2^{3})$      &  $(2^{-1}, 2^{5})$         &    $(2^{-2}, 2^{-8}, 2^{2})$        \\
ilpd\_indian\_liver               & 58.49    & 70.69  & 71.84 & 69.88  & \textbf{75.86}     & \underline{72.41}      \\
$(583 \times 9)$      & $(2^{-3}, 1, 2^{3})$         & $(2^{-2}, 2^{3})$       &   $(2^{-2}, 2^{-8}, 2^{-3})$ &  $(2^{-5}, 2^{-3}, 0.75, 2^{5})$     &   $(2^{-9}, 2^{-3})$        &   $(2^{-6}, 2^{-8}, 2^{2})$         \\
hepatitis                         & 67.58    & 63.04  & \underline{82.61} & \textbf{83.88}   & 76.09     & \underline{82.61}      \\
$(155 \times 19)$       & $(2^{-8}, 2^{-5}, 2^{3})$         & $(2, 1)$       &  $(2^{-3}, 2^{-8}, 2^{7})$  &  $(2^{5}, 2^{2}, 0.875, 2^{2})$     &  $(2^{3}, 2^{8})$         &     $(2^{-5}, 2^{-8}, 2^{2})$       \\
 \hline
 \multicolumn{7}{l}{$^{\dagger}$ represents the proposed model.}\\
 \multicolumn{7}{l}{Boldface and underline depict the best and second-best models, respectively.}
 \end{tabular}}
\end{table}

\begin{table}[htp]
    \centering
    \caption*{\text{Table 4: } (Continued)}
   \resizebox{0.9\textwidth}{!}{
    \begin{tabular}{lcccccc}
    \hline
\text{Model} $\rightarrow$ & Pin-GTSVM \cite{tanveer2019general}         & GEPSVM \cite{mangasarian2005multisurface}         & IGEPSVM \cite{shao2012improved}   & CGFTSVM-ID \cite{anuradha2024}        & IF-GEPSVM$^{\dagger}$        & IF-IGEPSVM$^{\dagger}$        \\ \hline
Dataset $\downarrow$     & $ACC$ (\%)              & $ACC$ (\%)  & $ACC$ (\%)  & $ACC$ (\%)       & $ACC$ (\%)              & $ACC$ (\%)              \\
     & $(c_1, c_2, \sigma)$     & $(\delta, \sigma)$   & $(\delta, \eta, \sigma)$ & $(c_1, c_2, r, \sigma)$    & $(\delta, \sigma)$     & $(\delta, \eta, \sigma)$     \\ \hline
hill\_valley                      & 67.76    & 65.89  & 69.87 & 56.49   & \underline{70.42}     & \textbf{71.89}      \\
$(1212 \times 100)$                                  & $(2^{-8}, 2^{-8}, 2^{-4})$         &  $(2^{-8}, 2^{2})$      & $(2^{4}, 2^{-7}, 2^{-4})$   &  $(2^{-4}, 2^{4}, 0.625, 2^{-5})$     &  $(2^{-4}, 2^{-7})$         &   $(2^{-8}, 2^{-2}, 2^{4})$         \\
iono                              & 88.89    & 85.71  & 81.84 & \underline{91.61}   & 76.19     & \textbf{91.84}      \\
$(351 \times 34)$   & $(2^{3}, 2^{-7}, 2^{-7})$         & $(2^{8}, 2^{-2})$       &  $(2^{-7}, 2^{-8}, 2^{5})$  &  $(2^{-5}, 2^{2}, 0.625, 2^{1})$     & $(2^{-4}, 2^{-7})$          &   $(2^{-7}, 2^{-8}, 2^{5})$         \\
ionosphere                        & 89.46    & \underline{90.48}  & 72.38 & \textbf{90.68}   & 86.67     & 88.57      \\
$(351 \times 33)$     &  $(2^{-8}, 2^{-4}, 2^{5})$        &  $(2^{6}, 2)$      & $(2^{-4}, 2^{-8}, 2^{4})$  &  $(2^{-7}, 2^{-1}, 0.5, 1)$      &  $(2^{-8}, 2^{2})$         &    $(2^{-3}, 2^{-8}, 2^{3})$        \\
magic                             & 54.56    & 45.89  & 52.43 & 58  & \textbf{62.67}     & \underline{58.87}      \\
$(19020 \times 10)$     & $(2^{-8}, 2^{-6}, 2^{-4})$         & $(2^{-4}, 2^{3})$       &  $(2^{-8}, 2^{3}, 2^{4})$  &  $(2^{-8}, 2^{-7}, 0.625, 2^{-7})$     &  $(2^{-8}, 2^{2})$         &    $(2^{-8}, 2^{-8}, 2^{-2})$        \\
monks\_1                          & \underline{79.77}    & 46.99  & 59.99  & 72.4  & 58.43     & \textbf{88.55}      \\
$(556 \times 6)$   & $(2^{-8}, 2, 2^{7})$         &  $(2, 2^{-8})$      &  $(2^{-5}, 2^{-8}, 2^{4})$  &  $(2^{-2}, 1, 0.5, 2^{4})$     &  $(2^{-8}, 2^{-1})$         &   $(2^{-8}, 2^{-8}, 2)$         \\
monks\_2                          & 71.96    & 75.56  & 73.33 & 79.39  & \underline{79.44}     & \textbf{86.67}      \\
$(601 \times 6)$    & $(2^{-3}, 2, 1)$         & $(2^{-8}, 2^{-8})$       & $(2^{-8}, 2, 2^{6})$   &  $(2^{-4}, 2^{2}, 0.5, 1)$     &  $(2^{-1}, 2^{-1})$         &    $(2^{-8}, 2^{-8}, 2^{2})$        \\
monks\_3                          & 76.99    & 47.59  & \underline{86.14}  & \textbf{93.98}  & 75.9      & 77.11      \\
$(554 \times 6)$  &$(1, 1, 2^{2})$          & $(2^{4}, 2^{-6})$       & $(2^{-8}, 2^{-8}, 2)$      &  $(2^{-5}, 2^{2}, 1, 2^{1})$  &  $(2, 2)$         &     $(2^{-7}, 2^{-8}, 2)$       \\                                  
mammographic                      & 68.03    & 75.35  & \textbf{82.64} & 82.05  & 71.88     & \underline{82.29}      \\
$(961 \times 5)$    & $(2^{-7}, 2^{-8}, 2^{6})$         & $(2^{-2}, 1)$       & $(2^{-3}, 2^{-8}, 2^{8})$  &  $(2^{-1}, 2^{-2}, 0.625, 2^{1})$      &  $(2^{8}, 2^{7})$         &     $(2^{-7}, 2^{-8}, 1)$       \\
molec\_biol\_promoter             & 86.75    & 48.39  & 70.97 & 61.54  & \underline{88.89}     & \textbf{93.55}      \\
$(106 \times 57)$   & $(2^{-2}, 2^{-5}, 2^{6})$         & $(2^{5}, 2^{7})$       & $(2^{3}, 2^{-8}, 2^{7})$ &  $(2^{2}, 2^{1}, 0.5, 2^{4})$       &  $(2^{-2}, 2^{5})$         &    $(2^{4}, 2^{-8}, 1)$        \\
mushroom                          & 65.64    & 65.67  & 68.79  & 70.49  & \underline{71.82}     & \textbf{75.65}      \\
$(8124 \times 21)$   & $(2^{-2}, 2^{3}, 2^{-4})$         & $(2^{-5}, 2^{6})$       & $(2^{8}, 2^{-5}, 2^{-4})$  &  $(2^{-8}, 2^{-8}, 0.5, 2^{-5})$      & $(2^{-4}, 2^{-7})$          &   $(2^{4}, 2^{-7}, 2^{-2})$         \\
oocytes\_merluccius\_nucleus\_4d  & 58.91    & 59.93  & 50.59  & \textbf{81.4} & \underline{79.86}     & 74.84      \\
$(1022 \times 41)$   &  $(2^{-1}, 2^{-3}, 2^{4})$        &  $(2^{-1}, 2^{-5})$      &  $(2^{-3}, 2^{-8}, 2^{-1})$  &  $(2^{-5}, 2^{1}, 0.75, 2^{2})$     &  $(2^{8}, 2^{5})$         &    $(2^{-5}, 2^{-8}, 2^{5})$        \\
oocytes\_trisopterus\_nucleus\_2f & 68.1     & 65.93  & 55.93 & \underline{78.53}   & 72.45     & \textbf{79.65}      \\
$(912 \times 25)$   &  $(2^{5}, 2, 2^{5})$        &  $(1, 2^{3})$      &  $(2^{-6}, 2^{-8}, 2^{7})$   &  $(2^{-5}, 2^{4}, 0.875, 2^{2})$    &  $(2^{2}, 1)$         &     $(2^{-3}, 2^{-8}, 2^{5})$       \\                                  
musk\_1                           & 69.2     & 64.54  & 60.56 & \textbf{89.36}  & \underline{74.15}     & 67.18      \\
$(476 \times 166)$   &  $(2^{8}, 2^{-5}, 2^{-5})$        &  $(2^{-7}, 2^{-2})$      &  $(2^{5}, 2^{-8}, 2^{6})$  &  $(2^{-2}, 1, 0.5, 2^{4})$     &  $(2^{6}, 2^{5})$         &   $(2^{2}, 2^{-8}, 2^{-1})$         \\
musk\_2                           & 84.67    & 82.32  & 79.8  & 85  & \underline{85.67}     & \textbf{89.94}      \\
$(6598 \times 166)$   & $(2^{8}, 2^{4}, 2^{3})$          & $(2^{8}, 2^{4})$       &  $(2^{6}, 2^{-7}, 2^{-4})$ &  $(2^{-4}, 2^{2}, 0.5, 1)$      &   $(2^{4}, 2^{7})$        &  $(2^{3}, 2^{-8}, 2^{-5})$          \\
ozone                             & 87.89    & 83.64  & 73.82 & 79.3  & \underline{92.45}     & \textbf{96.67}      \\
$(2536 \times 72)$    & $(2^{8}, 2^{-3}, 2^{-7})$         & $(2^{2}, 2^{-8})$       & $(2^{6}, 2^{-8}, 2^{2})$   &  $(2^{-8}, 2^{-8}, 0.5, 2^{-5})$     & $(2^{5}, 2^{-2})$          &  $(2^{-6}, 2^{-8}, 2^{-3})$          \\
parkinsons                        & 71.37    & 74.14  & 76.21 & 71.28  & \textbf{93.1}      & \underline{81.03}      \\
$(195 \times 22)$   & $(2^{7}, 2^{-4}, 2^{7})$         &  $(2^{8}, 2^{-1})$      &   $(2^{-1}, 2^{-8}, 2^{2})$ &  $(2^{3}, 2^{4}, 0.75, 2^{4})$     &   $(2^{5}, 2^{2})$        &    $(2^{3}, 2^{-8},1)$        \\
pima                              & 68.45    & 68.26  & \textbf{76.52} & \underline{73.81}  & 69.57     & 73.74      \\
$(768 \times 8)$     & $(2^{-4}, 2^{7}, 2^{3})$         & $(2^{5}, 2^{-4})$       &  $(1, 2^{-8}, 2^{2})$    &  $(2^{4}, 2^{-1}, 1, 2^{1})$   &   $(2^{-7}, 2^{5})$        &    $(2^{-1}, 2^{-8}, 2^{2})$        \\
pittsburg\_bridges\_T\_OR\_D      & 46.67    & \underline{83.33}  & 81.27 & 73.81  & 73.33     & \textbf{100}        \\
$(102 \times 7)$   & $(2^{-8}, 2^{-8}, 2^{5})$         & $(2^{-1}, 2^{-6})$       &$(2^{-5}, 2^{-8}, 2^{7})$   &  $(2^{2}, 2^{1}, 1, 2^{1})$      &  $(2^{-1}, 2^{6})$         &  $(2^{-7}, 2^{-8}, 2^{3})$          \\
planning                          & 45.83    & \textbf{75.93}  & 66.67 & 68.89  & \underline{72.22}     & 66.67      \\
$(182 \times 12)$    & $(2^{-3}, 2^{-8}, 2^{6})$         &  $(2^{-8}, 2^{-8})$      &$(2^{-4}, 2^{-8}, 2^{3})$  &  $(2^{-4}, 2^{1}, 1, 2^{4})$      &  $(2^{-8}, 2^{-8})$         &     $(2^{-8}, 2^{-8}, 2^{-8})$       \\
ringnorm                          & 64.45    & 67.89  & 58.83 & 71.58  & \underline{72.65}     & \textbf{80.54}      \\
$(7400 \times 20)$   & $(2^{-8}, 2^{5}, 2^{2})$         &  $(2^{-2}, 2^{2})$      & $(2^{-8}, 2^{-8}, 2^{-4})$  &  $(2^{-8}, 2^{-8}, 0.5, 2^{-5})$      &  $(2^{-4}, 2^{-7})$         &    $(2^{3}, 2^{-6}, 2^{5})$        \\
sonar                             & 69.89    & 54.84  & 64.83 & \textbf{81.9}  & \underline{70.95}     & 64.83      \\
 $(208 \times 61)$      & $(1, 2^{5}, 2^{7})$         &  $(2^{3}, 1)$      & $(2^{-3}, 2^{-8}, 2^{7})$  &  $(2^{-2}, 2^{-4}, 0.5, 2^{2})$      &  $(2^{-2}, 2)$         &      $(2^{-3}, 2^{-8}, 2^{7})$      \\
spambase                          & 78.65    & 76.87  & 65.42 & 85.45  & \underline{86.64}     & \textbf{87.42}      \\
$(4601 \times 57)$     & $(2^{-6}, 2^{-2}, 2^{-7})$         & $(2^{6}, 2^{8})$       & $(2^{2}, 2^{-8}, 2^{-4})$  &  $(2^{-5}, 2^{-4}, 0.5, 2^{-5})$     & $(2^{5}, 2^{-3})$          &    $(2^{-6}, 2^{-3}, 2^{-2})$        \\ 
statlog\_australian\_credit       & 45.14    & 68.12  & \underline{69.57} & 58.33  & 68.12     & \textbf{74.05}      \\
$(690 \times 14)$     & $(2^{8}, 2^{-4}, 2^{8})$         & $(2, 2^{5})$       &  $(2^{-8}, 2^{-8}, 1)$   &  $(1, 2^{-5}, 0.75, 2^{1})$    &   $(2^{-8}, 2^{-8})$        &     $(2^{-3}, 2^{-8}, 2^{-1})$       \\
statlog\_german\_credit           & 70.79    & 75     & 74.67 & \underline{75.47}  & 73.33     & \textbf{78.67}      \\
$(1000 \times 24)$    &  $(2^{-8}, 2^{-5}, 2^{7})$        & $(2^{7}, 2^{6})$       & $(2, 2^{-8}, 2^{4})$   &  $(2^{5}, 2^{2}, 1, 2^{3})$     &  $(2^{7}, 2^{7})$         &    $(2^{-4}, 2^{-8}, 2^{6})$        \\
spect                             & 63.1     & 55.7   & \textbf{67.09} & 65.69  & \underline{65.82}     & 59.49      \\
$(265 \times 22)$  & $(2^{-8}, 2^{-4}, 2^{4})$         &  $(2^{8}, 2^{5})$      &  $(2^{-2}, 2^{-8}, 2^{7})$ &  $(2^{-3}, 2^{-4}, 0.875, 2^{2})$      &  $(2^{5}, 2^{4})$         &   $(1, 2^{-8}, 2^{4})$         \\
spectf                            & 77.82    & 72.5  & 72.5  & \underline{85.85}  & 78.75      & \textbf{90}         \\
$(267 \times 44)$    & $(2^{-3}, 2^{-8}, 2^{-1})$         & $(2^{-8}, 2^{-8})$       & $(2^{-5}, 2^{-8}, 2^{-2})$  &  $(2^{-5}, 2^{-2}, 0.5, 2^{5})$      &  $(2^{-8}, 2^{-8})$         &    $(2^{-2}, 2^{-8}, 1)$        \\
statlog\_heart                    & 78.88    & 77.78  & \underline{82.72} & \textbf{84.56}   & 81.73     & \underline{82.72}      \\
$(270 \times 13)$  & $(2^{6}, 2^{6}, 2^{4})$         & $(2^{8}, 2^{8})$       & $(1, 2^{-8}, 2^{3})$  &  $(2^{-5}, 2^{-5}, 1, 2^{2})$      &  $(2^{-8}, 2^{3})$        &     $(2^{-6}, 2^{-8}, 2^{2})$       \\
titanic                           & 52.69    & 48.46  & 68.03 & 70.18  & \underline{76.36}     & \textbf{81.63}      \\
$(2201 \times 3)$    & $(2^{-8}, 2^{-7}, 1)$         & $(2^{-4}, 2^{-8})$       &  $(2^{-1}, 2^{-8}, 2^{2})$ &  $(2^{-1}, 1, 0.5, 2^{5})$      &  $(2, 2^{-1})$         &   $(2^{6}, 2^{-6}, 2^{2})$         \\
twonorm        & 65.57    & 69.85  & 59.63 & 75.78  & \underline{76.85}     & \textbf{79.89}      \\
$(7400 \times 20)$  &  $(2^{-7}, 2^{4}, 2^{6})$        &  $(2^{-4}, 2^{6})$      & $(2^{-6}, 2^{-4}, 2^{6})$  &  $(2^{-5}, 2^{3}, 0.5, 2^{-4})$      & $(2^{-7}, 2^{-2})$          &   $(2^{2}, 2^{-8}, 2^{-2})$         \\      
tic\_tac\_toe                     & 66.81    & 62.37  & 67.91 & \textbf{98.94}  & 72.47     & \underline{78.87}      \\
$(958 \times 9)$    & $(2^{-1}, 1, 2^{4})$         & $(2^{8}, 2^{-4})$       &  $(2^{-7}, 2^{-8}, 2^{8})$ &  $(2^{-5}, 2^{1}, 0.75, 2^{2})$      &  $(2^{6}, 2^{8})$         &   $(2^{-8}, 2^{-8}, 2^{-2})$         \\
vehicle1                          & 73.66    & 74.31  & 74.83  & \textbf{76.46} & \underline{74.95}     & 74.83      \\
$(846 \times 19)$   &  $(2^{-6}, 2^{3}, 2^{5})$        & $(2^{-6}, 2^{8})$       & $(2^{-8}, 2^{-8}, 2^{7})$    &  $(2^{-5}, 2^{-1}, 0.5, 2^{-1})$    &   $(2^{-8}, 2^{-5})$        &     $(2^{-8}, 2^{-8}, 2^{7})$       \\
vertebral\_column\_2clases        & 65.4     & 64.48  & 58.82 & 75.4  & \underline{75.48}     & \textbf{76.78}      \\
$(310 \times 6)$   & $(2^{6}, 2^{2}, 2^{4})$         &  $(2^{8}, 1)$      &  $(2^{3}, 2^{-8}, 2^{8})$ &  $(2^{-4}, 2^{4}, 0.875, 2^{2})$      &   $(2^{-3}, 2^{4})$        &    $(2^{7}, 2^{-8}, 2^{5})$        \\
vowel         & 93.20    & 91.27  & 93.20  & \underline{97.82}  & 95.27     & \textbf{99.82}       \\
$(988 \times 11)$     & $(2^{-3}, 2^{5}, 2^{4})$         &  $(2^{8}, 1)$      & $(2^{-8}, 2^{-8}, 2^{4})$   &  $(2^{5}, 2^{2}, 0.75, 2^{1})$    &   $(2^{-7}, 2^{2})$        &    $(2^{-8}, 2^{-8}, 2^{2})$        \\
wpbc                              & 62.87    & \underline{74.14}  & 60.37  & 61.01 & \textbf{75.86}     & 70.37      \\
$(194 \times 34)$  & $(1, 2^{3}, 2^{8})$         &  $(2^{2}, 2^{-2})$      &  $(2^{-5}, 2^{-8}, 2^{-7})$ &  $(2^{-2}, 1, 1, 2^{3})$      &   $(2^{-8}, 2^{-7})$        &    $(2^{-5}, 2^{-8}, 2^{7})$        \\ \hline
Average $ACC$     & 70.81    & 70.64  & 73.63 &  76.37   & \underline{77.98}     & \textbf{81.53}      \\ \hline
Average Rank        & 4.65  & 4.69   &  3.94  & 3.13   &  \underline{2.69} & \textbf{1.9}   \\ \hline
\multicolumn{7}{l}{$^{\dagger}$ represents the proposed model.}\\
 \multicolumn{7}{l}{Boldface and underline depict the best and second-best models, respectively.}
\end{tabular}}
\end{table}

Furthermore, to analyze the models, we use pairwise win-tie-loss sign test \cite{demvsar2006statistical}. As per the win-tie-loss sign test, the null hypothesis assumes that the two models are considered equivalent if each model wins approximately $N/2$ datasets out of the total $N$ datasets. At $5\%$ level of significance, the two models are considered significantly different if each model wins on approximately $\frac{N}{2} + 1.96\frac{\sqrt{N}}{2}$ datasets. If there is an even occurrence of ties between any two models, the ties are distributed equally among them. However, if the number of ties is odd, one tie is ignored, and the remaining ties are distributed equally among the given models. For $N = 62$, if one of the models’ wins is at least $38.716$ then there exists a significant difference between the models. Table \ref{Pairwise Win-tie using linear kernel} illustrates the comparative performance of the proposed IF-GEPSVM and IF-IGEPSVM models along with the baseline models, presenting their outcomes in terms of pairwise wins, ties, and losses using UCI and KEEL datasets. In Table \ref{Pairwise Win-tie using linear kernel}, the entry $[x, y, z]$ indicates that the model mentioned in the row wins $x$ times, ties $y$ times, and loses $z$ times in comparison to the model mentioned in the respective column. Table \ref{Pairwise Win-tie using linear kernel} distinctly illustrates that the proposed IF-IGEPSVM model demonstrates significant superiority compared to the baseline. Moreover, the proposed IF-GEPSVM model attains statistically significant distinctions from CGFTSVM-ID. Showcasing a notable level of performance, the proposed IF-GEPSVM model outperforms in $33$ out of $62$ datasets. Consequently, the proposed IF-GEPSVM and IF-IGEPSVM models exhibit significant superiority over existing models.

\begin{table}[htp]
    \centering
    \caption{Pairwise win-tie-loss test of proposed IF-GEPSVM and IF-IGEPSVM models along with baseline models on UCI and KEEL datasets with non-linear kernel.}
    \label{Pairwise Win-tie using Gaussian kernel}
   \resizebox{1.0\textwidth}{!}{
    \begin{tabular}{lcccccc}
    \hline
           & Pin-GTSVM \cite{tanveer2019general}       & GEPSVM \cite{mangasarian2005multisurface}          & IGEPSVM \cite{shao2012improved}  &  CGFTSVM-ID \cite{anuradha2024}  & IF-GEPSVM$^{\dagger}$       \\ \hline
GEPSVM \cite{mangasarian2005multisurface}    & {[}29, 0, 33{]} &       &   &        \\
IGEPSVM \cite{shao2012improved}   & {[}37, 1, 24{]} & {[}38, 1, 23{]} &                 &   \\
CGFTSVM-ID \cite{anuradha2024}   & {[}50, 0, 12{]}    & {[}45, 1, 16{]}   & {[}40, 0, 22{]}   &   &  \\
IF-GEPSVM$^{\dagger}$  & {[}53, 0, 9{]}  & {[}53, 2, 7{]}  & {[}43, 1, 18{]} &    {[}37, 0, 25{]}  &    \\
IF-IGEPSVM$^{\dagger}$ & {[}57, 0, 5{]}  & {[}58, 0, 4{]}  & {[}46, 9, 7{]} & {[}44, 1, 17{]}  & {[}43, 3, 16{]}   \\ \hline
\multicolumn{6}{l}{$^{\dagger}$ represents the proposed model.}
\end{tabular}}
\end{table}
For the non-linear case, the ACC values are shown in Table \ref{Average ACC of datasets using Gaussian kernel} for the proposed IF-GEPSVM and IF-IGEPSVM along with the baseline models. From Table \ref{Average ACC of datasets using Gaussian kernel}, it is evident that IF-GEPSVM and IF-IGEPSVM demonstrate superior generalization performance in most of the datasets. It is clear that the proposed IF-IGEPSVM and IF-GEPSVM secure the first and second position with an average ACC of $81.53\%$ and $77.98\%$, and the baseline models i.e., Pin-GTSVM, GEPSVM, IGEPSVM and CGFTSVM-ID has the average ACC of $70.81\%$, $70.64\%$, $73.63\%$, and $76.37\%$, respectively. The average ranks of all the models based on ACC values are shown in Table \ref{Average ACC of datasets using Gaussian kernel}. It can be noted that among all the models, our proposed IF-GEPSVM and IF-IGEPSVM hold the lowest average rank. The minimum rank of the model implies the better performance of the model. Furthermore, we conduct the Friedman test. For $N=62$ and $q=6$, we get $\chi_{F}^2 = 111.34$ and $F_{F}= 34.188$. Since $F_{F}=34.188 > 2.2435$, the null hypothesis is rejected, indicating a statistical distinction difference between the models. Next, the Nemenyi post hoc test is used to compare the models pairwise. At $5\%$ level of significance, the value of $C.D. = 0.9576$. 
The average rank differences between the proposed IF-GEPSVM and IF-IGEPSVM models with the baseline Pin-GTSVM, GEPSVM, IGEPSVM, and CGFTSVM-ID models are $(1.96, 2.75)$, $(2, 2.79)$, $(1.25, 2.04)$, and $(0.44, 1.23)$, respectively. As per the Nemenyi post hoc test, the proposed IF-GEPSVM and IF-IGEPSVM models exhibit significant differences compared to the baseline models, except for IF-GEPSVM with CGFTSVM-ID. However, the proposed IF-GEPSVM model surpasses the CGFTSVM-ID model in terms of average rank. Therefore, the proposed IF-GEPSVM and IF-IGEPSVM models demonstrated superior performance compared to the baseline models. We also perform a win-tie-loss sign test for the non-linear case. Table \ref{Pairwise Win-tie using Gaussian kernel} illustrates the pairwise win-tie-loss outcomes of the compared models on UCI and KEEL datasets. In our case, if either of the two models emerges victorious in at least $38.716$ datasets, they are considered statistically distinct. Table \ref{Pairwise Win-tie using Gaussian kernel} clearly illustrates that the proposed IF-GEPSVM and IF-IGEPSVM models exhibit statistically superior performance compared to the baseline Pin-GTSVM, GEPSVM, IGEPSVM, and CGFTSVM-ID models. Overall, from the preceding analysis, it is evident that the proposed IF-GEPSVM and IF-IGEPSVM models showcase competitive or even superior performance compared to the baseline models.

\begin{table}[ht!]
\centering
    \caption{Performance comparison of the proposed IF-GEPSVM and IF-IGEPSVM models against the baseline models on UCI and KEEL datasets with label noise.}
    \label{UCI and KEEL results with label noise}
    \resizebox{1.00\linewidth}{!}{
\begin{tabular}{lccccccc}
\hline
\text{Model} $\rightarrow$ & Noise & Pin-GTSVM \cite{tanveer2019general}        & GEPSVM \cite{mangasarian2005multisurface}         & IGEPSVM \cite{shao2012improved}  & CGFTSVM-ID \cite{anuradha2024}         & IF-GEPSVM$^{\dagger}$        & IF-IGEPSVM$^{\dagger}$       \\ \hline
Dataset $\downarrow$   &  & $ACC$ (\%)     & $ACC$ (\%)         & $ACC$ (\%)           & $ACC$ (\%)              & $ACC$ (\%)             & $ACC$ (\%)             \\ 
  &   & $(c_1, c_2)$     & $(\delta)$   & $(\delta, \eta)$  & $(c_1, c_2, r, \sigma)$   & $(\delta, \sigma)$     & $(\delta, \eta, \sigma)$      \\ \hline
acute\_nephritis & $5\%$ & \textbf{100} & 97.22 & 95.65 & \underline{97.83} & \textbf{100} & \textbf{100} \\
& & $(2^{-5}, 2^{-5})$ & $(2^{4})$ & $(2^{-8}, 2^{-8})$ & $(2^{-5}, 2^{-3}, 0.5, 2^{-5})$ & $(2^{1}, 2^{-8})$  &  $(2^{-8}, 2^{-1}, 2^{-8})$   \\
 & $10\%$ & \underline{97.83} & 63.89 & 90.85 & 91.3 & \textbf{100} & \textbf{100} \\
 & & $(2^{1}, 2^{-5})$ & $(2^{4})$ & $(2^{-8}, 2^{-8})$ & $(2^{-5}, 2^{-4}, 0.75, 2^{-5})$& $(2^{4}, 2^{-7})$ &  $(2^{-8}, 2^{4}, 2^{-5})$   \\
 & $15\%$ & 87.96 & \underline{94.44} & 90.79 & 89.13 & \underline{94.44} & \textbf{100} \\
 & & $(1, 2^{6})$ & $(2^{4})$ & $(2^{-8}, 2^{-8})$ & $(2^{2}, 2^{-5}, 0.5, 2^{-5})$ & $(2^{6}, 2^{-8})$ &  $(2^{-8}, 2^{4}, 2^{-8})$   \\
 & $20\%$ & 85.28 & \underline{97.22} & 89.45 & 93.48 & \textbf{100} & 83.33 \\ 
 & & $(2^{5}, 2^{-5})$ & $(2^{6})$ & $(2^{-8}, 2^{-8})$ & $(2^{3}, 2^{-4}, 0.5, 2^{3})$ & $(2^{-1}, 2^{-6})$ &  $(2^{-8}, 2^{-4}, 2^{-7})$   \\ \hline
Average ACC &  & 92.77 & 88.19 & 91.69 & 92.93 & \textbf{98.61} & \underline{95.83} \\ \hline
aus & $5\%$ & 87.56 & \underline{87.44} & 77.29 & 86.08 & \textbf{88.41} & \textbf{88.41} \\
& &$(2^{1}, 2^{-5})$ & $(2^{8})$ & $(2^{-8}, 2^{-8})$ & $(2^{-5}, 2^{-2}, 0.75, 2^{3})$ & $(2^{4}, 2^{-8})$ & $(2^{3}, 2^{6}, 2^{-8})$    \\
 & $10\%$ & 85.07 & 84.06 & 76.81 & \underline{86.34} & \textbf{88.41} & \textbf{88.41} \\
 & &$(2^{-5}, 2^{-2})$ &$(2^{8})$ & $(2^{-8}, 2^{-8})$ & $(2^{5}, 2^{2}, 1, 1)$ & $(2^{2}, 2^{-5})$ &  $(2^{-8}, 2^{2}, 2^{-5})$   \\
 & $15\%$ & 85.02 & 85.99 & 81.16 & \underline{86.88} & \textbf{88.41} & \textbf{88.41} \\
 & & $(2^{-1}, 2^{-3})$& $(2^{8})$ & $(2^{-8}, 2^{-8})$ & $(2^{4}, 2^{-5}, 1, 2^{-1})$ & $(2^{2}, 2^{-7})$& $(2^{-7}, 2^{2}, 2^{-8})$    \\
 & $20\%$ & \underline{88.54} & 85.02 & 73.43 & 84.25 & \textbf{89.37} & 80.19 \\ 
 & &$(2^{5}, 2^{-5})$ & $(2^{8})$ & $(2^{-8}, 2^{-8})$ & $(2^{5}, 2^{-1}, 0.75, 2^{4})$ & $(2^{2}, 2^{-6})$ &  $(2^{-8}, 2^{3}, 2^{-1})$   \\  \hline
Average ACC &  & \underline{86.55} & 85.63 & 77.17 & 85.89 & \textbf{88.65} & 86.35 \\ \hline
breast\_cancer\_wisc\_prog & $5\%$ & 68.87 & 59.32 & 76.27 & 71.81 & \underline{81.36} & \textbf{84.75} \\
& &$(2^{-5}, 2^{-4})$ & $(2^{5})$ & $(2^{-8}, 2^{-8})$ & $(2^{-3}, 2^{-1}, 0.75, 2^{3})$ & $(2^{4}, 2^{-6})$&  $(2^{-8}, 2^{2}, 2^{-6})$   \\  
 & $10\%$ & 66.91 & 55.93 & 77.97 & 68.87 & \underline{81.36} & \textbf{86.44} \\
 & & $(2^{-3}, 2^{-3})$ & $(2^{4})$ & $(2^{-8}, 2^{-8})$  & $(2^{-1}, 2^{-2}, 0.75, 2^{4})$ & $(2^{4}, 2^{-5})$ &  $(2^{-5}, 2^{-1}, 2^{-6})$   \\
 & $15\%$ & 57.6 & 57.63 & 71.19 & 57.35 & \underline{72.88} & \textbf{86.44} \\ 
 & &$(2^{2}, 2^{3})$ & $(2^{2})$ & $(2^{-7}, 2^{-8})$ &$(1, 2^{-5}, 0.875, 2^{4})$ & $(2^{4}, 2^{-6})$ &   $(2^{-3}, 2^{1}, 2^{-7})$  \\
 & $20\%$ & 46.69 & 59.32 & \textbf{74.58} & 45.59 & \textbf{74.58} & \underline{71.19} \\ 
 & &$(2^{7}, 2^{7})$ & $(2^{4})$ & $(2^{-7}, 2^{-8})$ & $(2^{4}, 2^{1}, 0.875, 2^{4})$ & $(2^{4}, 2^{-8})$ &   $(2^{-8}, 2^{6}, 2^{-8})$  \\  \hline
Average ACC &  & 60.02 & 58.05 & 75 & 60.91 & \underline{77.54} & \textbf{82.2} \\ \hline
credit\_approval & $5\%$ & 81.16 & 83.57 & \underline{86.47} & \textbf{86.95} & \underline{86.47} & 85.51 \\
& & $(2^{-5}, 2^{-3})$ & $(2^{-4})$ & $(2^{-8}, 2^{-8})$ & $(2^{-2}, 2^{3}, 0.5, 2^{1})$ & $(2^{2}, 2^{-7})$ &  $(2^{-8}, 2^{2}, 2^{-7})$   \\
 & $10\%$ & 76.17 & 82.13 & \underline{86.47} & \textbf{86.95} & 85.99 & 85.51 \\
 & & $(2^{-1}, 2^{-5})$ & $(2^{-7})$ & $(2^{-8}, 2^{-8})$ & $(1, 2^{-5}, 0.5, 2^{-5})$ & $(2^{2}, 2^{-6})$ & $(2^{-8}, 2^{2}, 2^{-3})$    \\
 & $15\%$ & 56.92 & 77.78 & \underline{85.99} & \textbf{86.95} & 85.51 & 85.51 \\
 & &$(2^{-5}, 2^{-3})$ & $(2^{-1})$& $(2^{-8}, 2^{-8})$ & $(2^{1}, 2^{-5}, 0.5, 2^{1})$ & $(2^{2}, 2^{-5})$ &  $(2^{-8}, 2^{4}, 2^{-6})$   \\ 
 & $20\%$ & \underline{85.73} & 77.29 & 81.64 & \textbf{86.95} & 85.51 & 85.51 \\ 
 & &$(1, 2^{-7})$ & $(2^{-7})$ & $(2^{-8}, 2^{-8})$ & $(2^{2}, 2^{-2}, 0.75, 2^{-5})$ & $(2^{2}, 2^{-3})$ &  $(2^{-8}, 2^{4}, 2^{-8})$   \\   \hline
Average ACC &  & 75 & 80.19 & 85.14 & \textbf{86.95} & \underline{85.87} & 85.51 \\ \hline
vehicle1 & $5\%$ & 63.45 & 67.98 & 74.31 & 75.36 & \textbf{78.66} & \underline{76.68} \\
& &$(2^{-4}, 2^{-7})$ & $(2^{-5})$ & $(2^{-8}, 2^{-8})$ & $(2^{3}, 2^{3}, 0.625, 2^{1})$ & $(2^{2}, 2^{-8})$ &   $(2^{-8}, 2^{4}, 2^{-2})$   \\
 & $10\%$ & 65.17 & 65.61 & \textbf{79.45} & 76.43 & \underline{77.08} & 75.89 \\
 & &$(2^{-3}, 2^{-1})$ & $(2^{-7})$ & $(2^{-8}, 2^{-8})$ & $(2^{1}, 1, 0.875, 2^{5})$&  $(2^{6}, 2^{-8})$&  $(2^{-2}, 2^{-4}, 2^{-8})$   \\
 & $15\%$ & 63.36 & 57.31 & 73.91 & \underline{75.21} & 72.33 & \textbf{75.89} \\  
 & & $(2^{-8}, 2^{-2})$ & $(2^{-5})$ & $(2^{-8}, 2^{-8})$ & $(2^{3}, 2^{1}, 0.75, 2^{5})$ & $(2^{3}, 2^{-7})$ &  $(2^{-8}, 2^{-4}, 2^{-8})$   \\
 & $20\%$ & 64.87 & 53.75 & \underline{75.49} & \textbf{79.27} & 73.91 & \underline{75.49} \\ 
 & & $(2^{-3}, 2^{2})$& $(2^{-5})$ & $(2^{-8}, 2^{-8})$ & $(1, 2^{-1}, 0.75, 2^{3})$ & $(2^{4}, 2^{2})$ &  $(2^{-7}, 2^{1}, 2^{-1})$   \\ \hline
Average ACC &  & 64.21 & 61.17 & 75.79 & \textbf{76.57} & 75.49 & \underline{75.99} \\ \hline
wpbc & $5\%$ & 67.52 & 70.69 & 75.86 & 67.36 & 74.14 & 74.14 \\
& &$(2^{-4}, 2^{2})$ & $(2^{-4})$ & $(2^{-8}, 2^{-8})$ & $(2^{2}, 1, 0.5, 2^{4})$ & $(2^{4}, 2^{-8})$ &   $(2^{-8}, 2^{2}, 2^{-8})$  \\
 & $10\%$ & 52.17 & 65.52 & \underline{70.69} & 63.02 & 62.07 & \textbf{74.14} \\ 
 & & $(2^{4}, 2^{5})$ & $(2^{2})$&  $(2^{-8}, 2^{-8})$ & $(2^{2}, 2^{1}, 0.75, 2^{4})$ & $(2^{6}, 2^{-8})$ &  $(2^{-8}, 2^{1}, 2^{-6})$   \\
 & $15\%$ & \textbf{82.02} & 65.52 & 65.52 & 59.53 & 65.52 & \underline{74.14} \\
 & &$(2^{-3}, 2^{4})$ & $(2^{-4})$ & $(2^{-8}, 2^{-8})$ & $(2^{1}, 2^{-5}, 0.875, 2^{3})$ & $(2^{8}, 2^{-4})$ & $(2^{-6}, 2^{1}, 2^{-4})$    \\ 
 & $20\%$ & 50.23 & 60.34 & \underline{72.41} & 66.2 & 62.07 & \textbf{75.86} \\  
 & &$(2^{6}, 2^{6})$ & $(2^{-7})$ & $(1, 2^{-8})$& $(1, 2^{-2}, 0.625, 2^{2})$ & $(2^{6}, 2^{-8})$ &  $(2^{-8}, 2^{6}, 2^{-7})$   \\  \hline
Average ACC &  & 62.98 & 65.52 & \underline{71.12} & 64.03 & 65.95 & \textbf{74.57} \\ \hline
Overall Average ACC  & & 73.59 & 73.13 & 79.32  & 77.88 & \underline{82.02}  & \textbf{83.41} \\ \hline
\multicolumn{7}{l}{$^{\dagger}$ represents the proposed model.}\\
 \multicolumn{7}{l}{Boldface and underline depict the best and second-best models, respectively.}
\end{tabular}}
\end{table}

\subsection{Evaluation on UCI and KEEL Datasets with Added Label Noise}
The evaluation conducted using UCI and KEEL datasets mirrors real-world scenarios. However, it's important to recognize that data impurities or noise may arise from various factors. To demonstrate the efficacy of the proposed IF-GEPSVM and IF-IGEPSVM models, particularly in challenging conditions, we intentionally introduced label noise to specific datasets. We chose $6$ datasets to test the robustness of the models namely acute\_nephritis, aus, breast\_cancer\_wisc\_prog, credit\_approval, vehicle1 and wpbc. To ensure fairness in model evaluation, we deliberately chose three datasets where the proposed IF-GEPSVM model did not attain the highest performance and three datasets where they achieved comparable results to an existing model with different levels of label noise. For a comprehensive analysis, we introduced label noise at different levels, including $5\%$, $10\%$, $15\%$, and $20\%$ to intentionally corrupt the labels of these datasets. Table \ref{UCI and KEEL results with label noise} presents the ACC of all models for the selected datasets with $5\%$, $10\%$, $15\%$, and $20\%$ noise. Consistently, the proposed IF-GEPSVM and IF-IGEPSVM models exhibit superior performance over baseline models, demonstrating higher ACC. Significantly, they sustain this leading performance despite the presence of noise. The average ACC of the proposed IF-GEPSVM and IF-IGEPSVM on the acute\_nephritis dataset at various noise levels are $98.61\%$ and $95.83\%$, respectively, surpassing the performance of the baseline models. On the aus dataset, both proposed models' ACC are lower than the CGFTSVM-ID at $0\%$ noise level (refer to Table \ref{Average ACC of datasets using linear kernel}). However, the average ACC of the proposed models at different noise levels are $88.65\%$ and $86.35\%$, respectively, outperforming all the baseline models. On the credit\_approval and vehicle1 datasets, the proposed models did not secure the top positions at $0\%$ noise level. However, with distinct noise levels, the proposed IF-GEPSVM and IF-IGEPSVM secured the second and third positions, respectively, on the credit\_approval and vehicle1 datasets. On different levels of noise, the proposed IF-GEPSVM and IF-IGEPSVM models, with an average ACC of $65.95\%$, and $74.57\%$, surpass all the baseline models on wpbc dataset. At each noise level, IF-GEPSVM and IF-IGEPSVM emerge as the top performers, with overall average ACC of $82.02\%$ and $83.41\%$, respectively. By subjecting the model to rigorous conditions, we aim to demonstrate the exceptional performance and superiority of the proposed IF-GEPSVM and IF-IGEPSVM models, particularly in unfavorable scenarios. The above findings emphasize the importance of the proposed IF-GEPSVM and IF-IGEPSVM models as resilient solutions, capable of performing well in demanding conditions marked by noise and impurities.

\subsection{Sensitivity Analysis}
In this subsection, we perform sensitivity analyses on the hyperparameters $\delta$ and $\sigma$. Additionally, we investigate the impact of varying levels of label noise on the model.

\begin{figure*}[htp]
\begin{minipage}{.30\linewidth}
\centering
\subfloat[acute\_nephritis \\ (IF-GEPSVM)]{\label{fig2:1a}\includegraphics[scale=0.25]{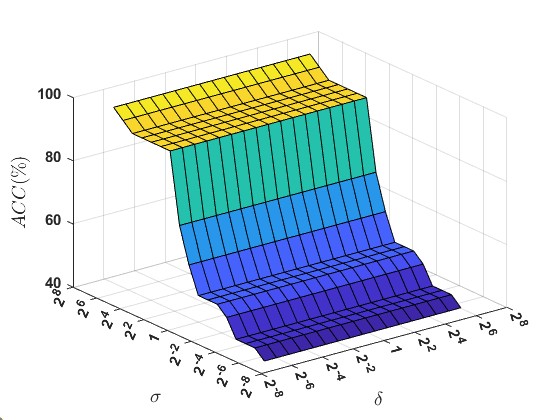}}
\end{minipage}
\begin{minipage}{.30\linewidth}
\centering
\subfloat[breast\_cancer\_wisc \\ (IF-GEPSVM)]{\label{fig2:1b}\includegraphics[scale=0.25]{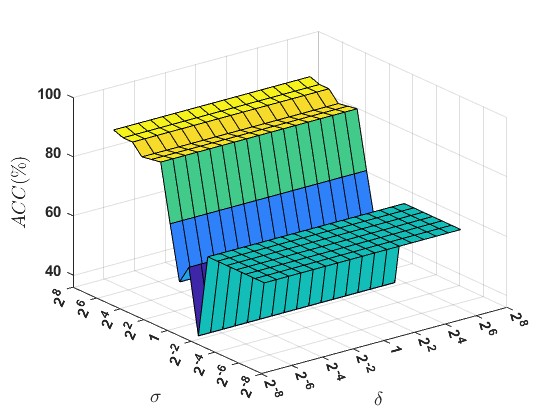}}
\end{minipage}
\begin{minipage}{.30\linewidth}
\centering
\subfloat[credit\_approval \\ (IF-GEPSVM)]{\label{fig2:1c}\includegraphics[scale=0.25]{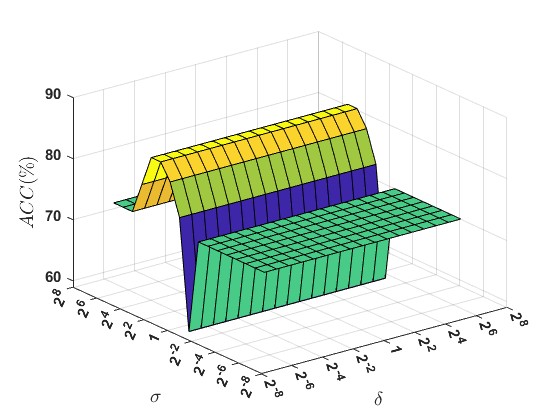}}
\end{minipage}
\par\medskip
\begin{minipage}{.30\linewidth}
\centering
\subfloat[acute\_nephritis \\ (IF-IGEPSVM)]{\label{fig2:11d}\includegraphics[scale=0.25]{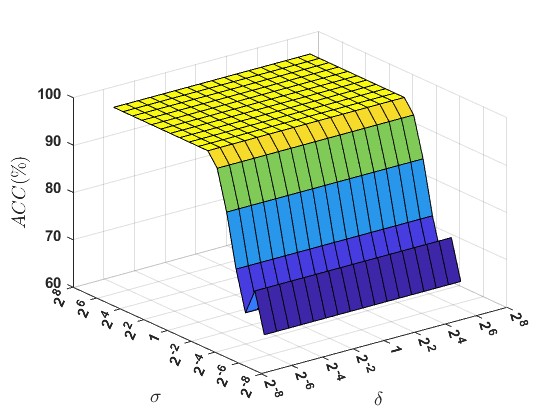}}
\end{minipage}
\begin{minipage}{.30\linewidth}
\centering
\subfloat[breast\_cancer\_wisc \\ (IF-IGEPSVM)]{\label{fig2:1e}\includegraphics[scale=0.25]{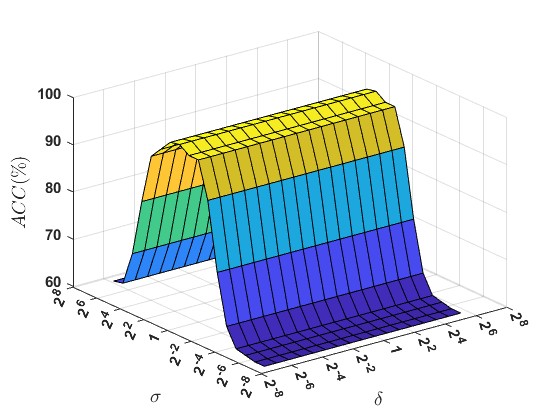}}
\end{minipage}
\begin{minipage}{.30\linewidth}
\centering
\subfloat[credit\_approval \\ (IF-IGEPSVM)]{\label{fig2:1f}\includegraphics[scale=0.25]{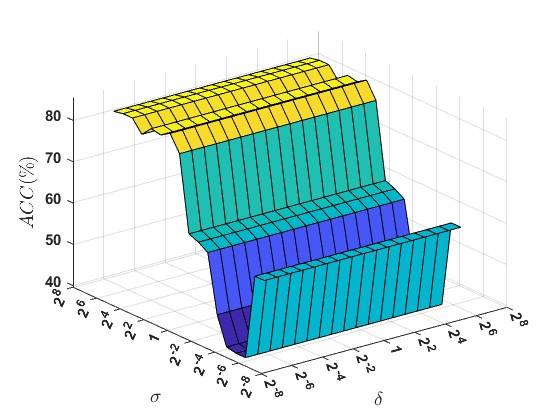}}
\end{minipage}
\caption{Effect of parameters $\delta$ and $\sigma$ on the performance of the proposed IF-GEPSVM and IF-IGEPSVM models.}
\label{Effect of parameters}
\end{figure*}

\subsubsection{Sensitivity Analysis of Hyperparameter \texorpdfstring{$\delta$}{delta} and \texorpdfstring{$\sigma$}{sigma}}
We assess the performance of the proposed IF-GEPSVM and IF-IGEPSVM models by adjusting the values of $\delta$ and $\sigma$. This comprehensive examination allows us to identify the configuration that optimizes predictive ACC and enhances the model's robustness against unseen data. Fig. \ref{Effect of parameters} depicts noticeable variations in the model's ACC across different combinations of $\delta$ and $\sigma$ values, highlighting the sensitivity of our model's performance to these particular hyperparameters. Based on the results depicted in Figs. \ref{fig2:1a} and \ref{fig2:11d}, we observe the optimal performance of the proposed IF-GEPSVM and IF-IGEPSVM models within the $\sigma$ ranges of $2^{4}$ to $2^8$ and $2^{2}$ to $2^8$. Figs. \ref{fig2:1b} and \ref{fig2:1e} demonstrates an increase in testing ACC of the proposed IF-GEPSVM and IF-IGEPSVM models within the $\sigma$ range spanning from $2^{2}$ to $2^8$ and $2^{-2}$ to $2^2$. Similarly, in Figs. \ref{fig2:1c} and \ref{fig2:1f}, the testing ACC within the ranges $1$ to $2^7$ and $2^{2}$ to $2^8$ of both the proposed IF-GEPSVM and IF-IGEPSVM models. These results suggest that, when considering the parameters $\sigma$ and $\delta$, the performance of both the proposed IF-GEPSVM and IF-IGEPSVM models is predominantly influenced by $\sigma$ rather than $\delta$. This highlights the importance of the kernel space and the effective extraction of non-linear features in the proposed IF-GEPSVM and IF-IGEPSVM models. Therefore, it is advisable to carefully consider the selection of the hyperparameter $\sigma$ in IF-GEPSVM and IF-IGEPSVM models to achieve superior generalization performance.

\begin{figure*}[ht!]
\begin{minipage}{.48\linewidth}
\centering
\subfloat[acute\_nephritis]{\label{fig:1a}\includegraphics[scale=0.36]{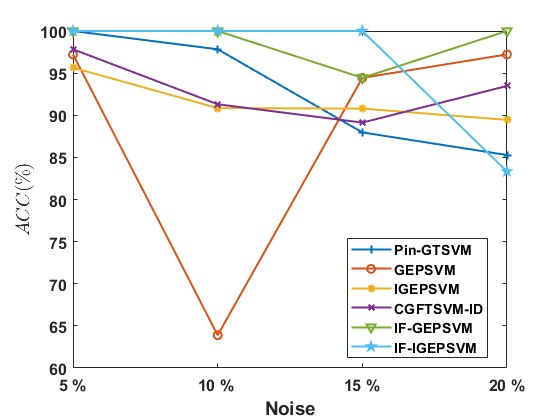}}
\end{minipage}
\begin{minipage}{.48\linewidth}
\centering
\subfloat[aus]{\label{fig:1b}\includegraphics[scale=0.36]{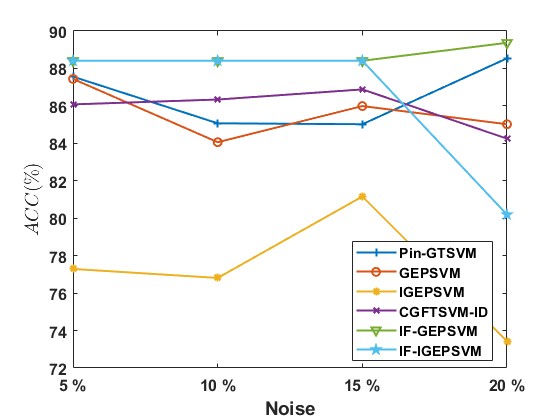}}
\end{minipage}
\par\medskip
\begin{minipage}{.48\linewidth}
\centering
\subfloat[credit\_approval]{\label{fig:1d}\includegraphics[scale=0.36]{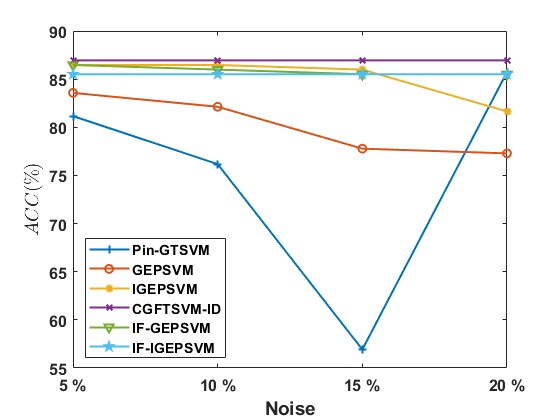}}
\end{minipage}
\begin{minipage}{.48\linewidth}
\centering
\subfloat[vehicle1]{\label{fig:1e}\includegraphics[scale=0.36]{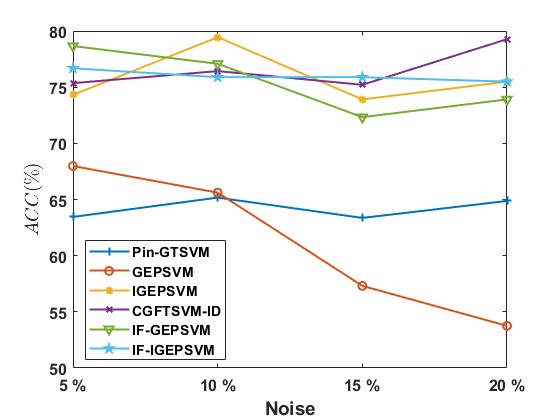}}
\end{minipage}
\caption{Effect of different labels of noise on the performance of the proposed IF-GEPSVM and IF-IGEPSVM models.}
\label{Effect of different labels of noise on the performance of the proposed IF-GEPSVM and IF-IGEPSVM models}
\end{figure*}

\subsubsection{Sensitivity Analysis of Label Noise}
The proposed IF-GEPSVM and IF-IGEPSVM models are designed with a primary focus on mitigating the adverse impact of noise. Their robustness is evidenced by their performance across different levels of label noise. The testing ACC of both the proposed IF-GEPSVM and IF-IGEPSVM models and the baseline models under varying levels of label noise is illustrated in Fig. \ref{Effect of different labels of noise on the performance of the proposed IF-GEPSVM and IF-IGEPSVM models}. The performance is shown for the acute\_nephritis, aus, credit\_approval, and vehicle1 datasets. The performance of baseline models exhibits notable fluctuations, diminishing notably with varying levels of noise labels. In contrast, the proposed IF-GEPSVM and IF-IGEPSVM models consistently maintain superior performance across different noise levels.

\begin{figure}[ht!]
\centering
     { %
\includegraphics[width=0.6\textwidth,height=4cm]{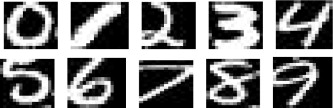}}
\caption{A visual representation of USPS database. }
    \label{USPS}
\end{figure}

\begin{table}[ht!]
    \centering
    \caption{Classification performance of proposed IF-GEPSVM and IF-IGEPSVM models and baseline models on USPS recognition datasets.}
    \label{classification results of USPS recognition dataset}
   \resizebox{1\textwidth}{!}{
    \begin{tabular}{lcccccc}
    \hline
 \text{Model} $\rightarrow$ & Pin-GTSVM \cite{tanveer2019general}        & GEPSVM \cite{mangasarian2005multisurface}         & IGEPSVM \cite{shao2012improved}  & CGFTSVM-ID \cite{anuradha2024}        & IF-GEPSVM$^{\dagger}$       & IF-IGEPSVM$^{\dagger}$       \\ \hline
Dataset $\downarrow$     & $ACC$ (\%)              & $ACC$ (\%)            & $ACC$ (\%)      & $ACC$ (\%)       & $ACC$ (\%)             & $ACC$ (\%)              \\
     & $(c_1, c_2)$     & $(\delta)$   & $(\delta, \eta)$  & $(c_1, c_2, r, \sigma)$    & $(\delta)$     & $(\delta, \eta)$      \\ \hline
0 vs. 1  & 88.65            & 53.78    & 88.35    &85.87      & \textbf{99.53}          & \underline{98.84}    \\
         & $(2^{-7}, 2^{-8})$ & $(2^{5})$    & $(2^{-6}, 2^{-8})$ & $(2^{5}, 2^{-3}, 1, 2^{-5})$ & $(2^{-4})$ & $(2^{-7}, 2^{-8})$ \\
1 vs. 2  & 89.34      & 42.49 & 98.18  & 95.69         & \textbf{99.09}    & \underline{98.57}    \\
         & $(2^{-8}, 2^{-5})$ & $(2^{-1})$ & $(2^{-5}, 2^{-8})$ & $(2^{3}, 2^{-3}, 0.875, 2^{-5})$ & $(2^{-7})$ & $(2^{-8}, 2^{-8})$ \\
2 vs. 3  & 87.5    & 95.24    & 92.19 & \underline{95.5}       & \textbf{96.38}   & 88.51      \\
         & $(2^{-8}, 2^{-4})$ & $(2^{4})$    & $(2^{-5}, 2^{-8})$ & $(1, 2^{-4}, 0.5, 2^{-5})$ & $(2^{2})$  & $(2^{-5}, 2^{-8})$ \\
3 vs. 4  & 89.38  & 91.04  & 89.84 & \underline{96.23}      & \textbf{98.01}   & 94.46    \\
         & $(2^{-8}, 2^{-8})$ & $(2^{8})$    & $(2^{-4}, 2^{-8})$ & $(2^{5}, 2^{-5}, 0.5, 2^{5})$ & $(2^{4})$  & $(2^{-5}, 2^{-8})$ \\
5 vs. 6  & 80.96   & 92.04   & \textbf{94.84} & 90.27    & 92.47     & \underline{93.82}    \\
         & $(2^{-7}, 2^{-8})$ & $(2^{7})$    & $(2^{-6}, 2^{-8})$ & $(2^{5}, 2^{-5}, 0.5, 2^{5})$ & $(2^{4})$  & $(2^{-7}, 2^{-8})$ \\
2 vs. 7  & 92.25 & \underline{97.67}     & 92.64 & 97.54  & \textbf{98.26}   & 96.1    \\
         & $(2^{-5}, 2^{-6})$ & $(2^{4})$    & $(2^{-5}, 2^{-8})$ & $(2^{5}, 2^{-4}, 0.5, 2^{4})$ & $(2^{1})$  & $(2^{-6}, 2^{-8})$ \\
3 vs. 8  & \underline{88.75}     & 85.19      & 88.67 & 87.29  & 87.8  & \textbf{88.90}  \\
         & $(2^{-8}, 2^{-2})$ & $(2^{8})$    & $(2^{-4}, 2^{-8})$ & $(2^{5}, 2^{-2}, 0.5, 2^{5})$ & $(2^{6})$  & $(2^{-4}, 2^{-8})$ \\
2 vs. 5  & \underline{91.99}    & 91.08    & 91.08 & 90.22   & \textbf{92.09}     & 87.92    \\
         & $(2^{-8}, 2^{-3})$ & $(2^{4})$    & $(2^{-4}, 2^{-8})$ & $(2^{5}, 2^{1}, 0.75, 2^{4})$ & $(2^{6})$  &$ (2^{-3}, 2^{-8})$ \\ \hline
Average ACC & 88.60     & 81.07    & 91.97 & 92.33    & \textbf{95.45}  & \underline{93.39}  \\ \hline 
\multicolumn{7}{l}{$^{\dagger}$ represents the proposed model.}\\
 \multicolumn{7}{l}{Boldface and underline depict the best and second-best models, respectively.}
\end{tabular} }
\end{table}

\subsection{USPS recognition}
The collection of grayscale images of handwritten digits ranging from $0$ to $9$ can be found in the USPS database, accessible at \url{https://cs.nyu.edu/~roweis/data.html}. This is depicted in Fig. \ref{USPS}. There are a total of $1100$ images for each digit in the dataset, and each image has a size of $16\times16$ pixels with $256$ different shades of gray (resulting in $1100$ examples per class with $256$ dimensions). In this case, we have chosen eight pairs of digits with varying levels of difficulty for classifying odd and even digits. The specific classes that have been selected can be found in Table \ref{classification results of USPS recognition dataset}. We use linear kernel to evaluate the experiment. The ACC of each model for classification is evaluated using the $10$-fold cross-validation. The performance of the proposed IF-GEPSVM and IF-IGEPSVM models along with the baseline models are presented in Table \ref{classification results of USPS recognition dataset}. Notably, IF-GEPSVM and IF-IGEPSVM models attained the first and second positions with average ACC of $95.45\%$ and $93.39\%$, respectively. In contrast, the baseline models, comprising Pin-GTSVM, GEPSVM, IGEPSVM, and CGFTSVM-ID, exhibited lower average ACC of $88.60\%$, $81.07\%$, $91.97\%$ and $92.33\%$, respectively. Compared to the third-top model, CGFTSVM-ID, the proposed IF-GEPSVM and IF-IGEPSVM models exhibit average ACC that are approximately $3.12\%$ and $1.06\%$ higher, respectively. Hence, the proposed IF-GEPSVM and IF-IGEPSVM models outperform in terms of ACC when compared to other baseline models.

\section{Conclusion}
\label{Conclusion}
In this paper, we proposed novel intuitionistic fuzzy generalized eigenvalue proximal support vector machine (IF-GEPSVM) model to reduce the effect of noise and outliers. The proposed IF-GEPSVM obtains two nonparallel hyperplanes by solving eigenvalue problems instead of QPP as in SVM. The classification of an input sample takes into account its membership and non-membership values, which aids in reducing the impact of noise and outliers. IF-GEPSVM remains susceptible to risks, such as the potential occurrence of singularity issues during the implementation of generalized eigenvalue decomposition. Furthermore, we propose a novel intuitionistic fuzzy improved generalized eigenvalue proximal support vector machine (IF-IGEPSVM). IF-IGEPSVM leverages intuitionistic fuzzy theory similar to the IF-GEPSVM. To demonstrate the effectiveness and efficiency of the proposed IF-GEPSVM and IF-IGEPSVM models, we conducted experiments on artificial datasets and $62$ UCI and KEEL datasets. The experimental results indicate that the proposed IF-GEPSVM and IF-IGEPSVM models beat baseline models in efficiency and generalization performance. The statistical measures based on the Friedman test and Nemenyi post-hoc test at a $5\%$ significance level deduce the coherence of the IF-GEPSVM and IF-IGEPSVM models over the baseline models. To assess the resilience of the proposed IF-GEPSVM and IF-IGEPSVM models, we introduced label noise to six diverse UCI and KEEL datasets. The proposed models demonstrated superior performance, effectively handling challenges posed by noise and impurities. To demonstrate the practical applications of the proposed IF-GEPSVM and IF-IGEPSVM models, we conducted experiments on the USPS recognition dataset. Experimental evaluation demonstrates the efficacy of the proposed IF-GEPSVM and IF-IGEPSVM models. It is perceptible that the proposed IF-GEPSVM and IF-IGEPSVM models are efficient in terms of ACC. Overall, IF-GEPSVM and IF-IGEPSVM are able to produce astonishing results. Extending the IF-GEPSVM and IF-IGEPSVM models to handle multi-category pattern classification and regression tasks would be an intriguing avenue for future research. The proposed models can be extended to real-world classification tasks, including speech recognition, natural language processing, and image segmentation, which commonly involve datasets with outliers and noise. The source code link of the proposed IF-GEPSVM and IF-IGEPSVM models are available at \url{https://github.com/mtanveer1/IF-GEPSVM}.

\section*{Acknowledgment}
This work is supported by Indian government's Department of Science and Technology (DST) through the MTR/2021/000787 grant as part of the Mathematical Research Impact-Centric Support (MATRICS) scheme.
\bibliography{refs1.bib}
\bibliographystyle{plainnat}
\end{document}